%% file: iclr2027_conference.tex
\pdfoutput=1

\documentclass{article}
\usepackage{iclr2027_conference,times}

\input{math_commands.tex}

\usepackage[table]{xcolor}
\usepackage{graphicx}
\usepackage{booktabs}
\usepackage{caption}
\usepackage{array}
\usepackage{multirow}
\usepackage{relsize}
\usepackage{algorithm}
\usepackage{algpseudocode}
\usepackage{hyperref}
\usepackage{url}
\usepackage{cleveref}
\crefname{figure}{Figure}{Figures}
\crefname{table}{Table}{Tables}
\crefname{section}{Section}{Sections}
\crefname{subsection}{Section}{Sections}
\crefname{subsubsection}{Section}{Sections}
\crefname{appendix}{Appendix}{Appendices}
\crefname{equation}{Equation}{Equations}
\crefname{algorithm}{Algorithm}{Algorithms}
\usepackage{arydshln}
\ADLinactivate
\usepackage{tcolorbox}
\tcbuselibrary{breakable}

\newcommand{\modelname}{\mbox{\textsc{CorpusMap}}}
\usepackage{pifont}
\definecolor{docfill}{HTML}{EAF3FC}\definecolor{docedge}{HTML}{9CC3E6}
\definecolor{entfill}{HTML}{FDEBD3}\definecolor{entedge}{HTML}{EDB46A}
\definecolor{walkfill}{HTML}{F0F8EC}\definecolor{walktext}{HTML}{286E2C}
\definecolor{failtext}{HTML}{B42318}
\definecolor{oursrow}{HTML}{E3F0FB}
\newcommand{\docchip}[1]{{\setlength{\fboxsep}{1pt}\fcolorbox{docedge}{docfill}{#1}}}
\newcommand{\entchip}[1]{{\setlength{\fboxsep}{1pt}\fcolorbox{entedge}{entfill}{#1}}}
\newcommand{\greencheck}{{\color{walktext}\ding{51}}}
\newcommand{\redx}{{\color{failtext}\ding{55}}}

\definecolor{linkblue}{RGB}{0,65,145}
\hypersetup{
 colorlinks=true,
 citecolor=linkblue,
 linkcolor=linkblue,
 urlcolor=linkblue
}

\title{Follow the Entities:\\A Corpus Map for Agentic Search}

\author{\mbox{Soyeong Jeong$^{1}$\thanks{Work done during an internship at Microsoft.}\hphantom{$^{*}$}}\hspace{0.6em}Sujay Kumar Jauhar$^{2}$\hspace{0.6em}Sung Ju Hwang$^{1}$\hspace{0.6em}Andrew Joohun Nam$^{2}$ \\[0.25em]
$^{1}$KAIST \quad $^{2}$Microsoft \\[0.25em]
\texttt{\fontsize{8.5pt}{8.5pt}\selectfont \{starsuzi, sungju.hwang\}@kaist.ac.kr, \{sjauhar, andrewnam\}@microsoft.com}}

\iclrfinalcopy
\begin{document}

\maketitle
\lhead{Preprint}

\input{Sections/1_abstract}
\input{Figures/fig_enterprise_rag_correctness}
\input{Sections/2_introduction}
\input{Sections/3_related_work}
\input{Sections/4_method}
\input{Sections/5_experimental_setup}
\input{Sections/6_experimental_results}

\input{Sections/7_conclusion}
\input{Sections/8_ai_use_statement}
\input{Sections/9_ethics_statement}
\input{Sections/10_reproducibility}

\bibliography{iclr2027_conference}
\bibliographystyle{iclr2027_conference}

\clearpage
\appendix
\input{Sections/13_appendix}

\end{document}

%% file: math_commands.tex
\usepackage{amsmath,amsfonts,bm}

\def\eqref#1{equation~\ref{#1}}

\def\1{\bm{1}}

\DeclareMathAlphabet{\mathsfit}{\encodingdefault}{\sfdefault}{m}{sl}
\SetMathAlphabet{\mathsfit}{bold}{\encodingdefault}{\sfdefault}{bx}{n}

\def\gC{{\mathcal{C}}}
\def\gD{{\mathcal{D}}}
\def\gE{{\mathcal{E}}}

\def\gG{{\mathcal{G}}}

\def\gL{{\mathcal{L}}}

\def\gN{{\mathcal{N}}}

\def\gR{{\mathcal{R}}}

%% file: Sections/1_abstract.tex
\begin{abstract}
Answering questions and completing tasks over large document collections often requires connecting evidence spread across multiple documents, such as a project's approval recorded in one, its requirements in another, and its latest status in a third.
Recent LLM agents approach this by iteratively searching the full corpus rather than reading only a fixed set of top-ranked documents.
However, when the corpus is exposed only as a flat collection of files, a relevant document gives no indication of how it relates to others, so the agent must rediscover these relationships for every query, often missing complementary evidence while simultaneously consuming substantial additional tokens.
To address this, we introduce \textsc{CorpusMap}, a navigation layer that organizes the corpus around its recurring entities, which are identifiable from the documents themselves and can link a single document to many others across sources.
Specifically, \textsc{CorpusMap} represents each recurring entity as an Entity Page that aggregates information about it and links to every document that refers to it, forming a graph between entities and documents that the agent can traverse to gather otherwise disconnected evidence.
Moreover, since \textsc{CorpusMap} is constructed offline by resolving mentions of the same entity across documents, its links are shared across queries rather than rediscovered repeatedly at inference time.
Using 7 different models with 3 benchmark datasets, we show that \textsc{CorpusMap} improves both evidence discovery and answer quality over raw-corpus agentic search while using fewer tokens on average, and further outperforms 4 alternative navigation layers, suggesting that entities serve as effective anchors for navigating large document collections.
\end{abstract}

%% file: Figures/fig_enterprise_rag_correctness.tex
\begingroup
\setlength{\intextsep}{6pt}
\begin{figure}[H]
    \centering
    \includegraphics[width=0.9\linewidth]{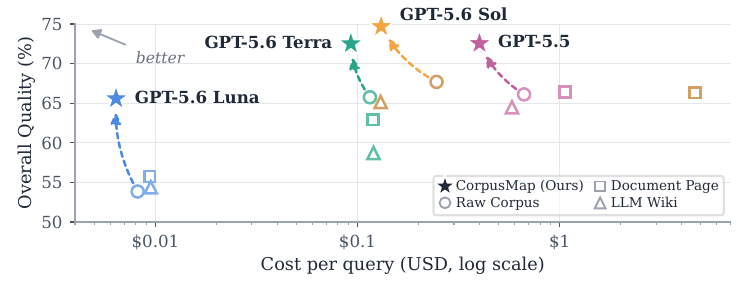}
    \vspace{-0.1in}
    \caption{Overall Quality (Table~\ref{tab:main_results}) versus estimated answer-generation cost for \modelname{} and the three strongest baselines, averaged equally over all benchmarks.}
    \label{fig:enterprise_rag_correctness}
\end{figure}
\endgroup

%% file: Sections/2_introduction.tex
\section{Introduction}
\label{sec:introduction}

Large language models (LLMs) have shown impressive capabilities as agents~\citep{gpt55-system-card, gpt56-system-card, DeepSeek-V4, mai-thinking-1}, and have been widely adopted to answer questions and complete tasks over large document collections~\citep{llm-agent-survey, deep-research-survey, agentic-rag-survey1}, where the necessary evidence is often distributed across multiple documents.
For example, determining whether a project is ready to launch may require combining its latest status from a project tracker, an approval recorded over email, requirements from operational documents, and a final decision recorded in meeting notes.
Although each source captures part of the answer, none is sufficient in isolation, and the relationships among them may not be stated explicitly in any individual document.
Moreover, in practice, these sources are buried among hundreds of thousands of other documents spread across different applications (e.g., issue trackers, shared drives, and chat channels)~\citep{enterpriseragbench, HERB}, so that the challenge lies not only in utilizing them, but also in locating and retrieving them.

To locate supporting evidence, retrieval-augmented generation (RAG) typically retrieves documents relevant to a user query or instruction~\citep{rag, bm25, dpr, rag_survey1} by selecting the top-$k$ documents before starting the model's reasoning or response. 
For tasks that require follow-up searches for additional documents, agentic search retrieves evidence over multiple steps~\citep{react, agentic-rag-survey1} and uses tool calls to search the full corpus directly~\citep{keyword-search-agent, dci}, so that it is no longer limited to what an initial retrieval step returns.
Nevertheless, full-corpus access does not by itself make the corpus easy to navigate, since it remains a flat collection of documents, without representing their relationships.

As a result, the agent is left to infer these relationships on its own by searching, reading, and reasoning over multiple sources sequentially, limiting both efficacy and efficiency.
For instance, a document's relevance to a query may be opaque or require corpus-specific knowledge (e.g. meeting notes that record the launch decision under the project's internal codename), so that searching with the query alone can miss it, despite being accessible.
Moreover, because each query is treated independently, the agent must spend a substantial number of tokens to re-read and re-discover these relationships each time.
Yet, while different queries require different subsets of these relationships, the relationships themselves (e.g. which documents refer to the same project) remain stable across queries, and could thus be identified in advance and reused.
We therefore frame this challenge as a \emph{corpus navigation} problem, which calls for a persistent \emph{navigation layer} that exposes reusable cross-document structure while preserving access to the full corpus.

This raises a central design question: which relationships should such a layer expose?
Given the challenges above, they should link documents that the query text alone may not reach easily and be identifiable in advance so that they can be reused across queries.
In our work, we leverage the fact that documents are naturally generated around common entities, such as people, projects, or products, and design a navigation layer that makes these entities and their related documents explicit.
Since entities and their relations are identifiable by the corpus alone (and not any queries), the navigation layer can be built entirely offline, so that query-time inference remains efficient.

\input{Figures/fig_concept}

To this end, we introduce \modelname{}, a novel \emph{entity-centric} navigation layer that makes these links explicit and reusable.
\modelname{} is constructed offline by identifying entity mentions within each document, resolving those that refer to the same entity across sources, and representing each resolved entity as an \emph{Entity Page} that gathers what different documents state about it, attributing each fact to its source, and links to every document that refers to it.
Importantly, Entity Pages add a layer over the corpus rather than replacing it, so that the original documents remain available to the model. 
As illustrated in \cref{fig:concept}, from any document it reads, the agent can follow the mentioned entities to complementary evidence instead of searching for it again, which can surface otherwise overlooked evidence while reducing the documents to inspect.

We validate \modelname{} across multiple LLMs on complex questions spanning multiple documents from three benchmarks, EnterpriseRAG-Bench~\citep{enterpriseragbench}, WixQA~\citep{WixQA}, and HERB~\citep{HERB}, and find that it consistently improves both evidence discovery and answer quality over raw-corpus agentic search and alternative navigation layers (e.g., LLM Wiki~\citep{llm-wiki} and Corpus2Skill~\citep{corpus2skill}).
Specifically, compared with raw-corpus agentic search, \modelname{} improves overall quality by 6.4 to 11.7 points while using 34\% to 57\% fewer input tokens on average, as summarized in \cref{fig:enterprise_rag_correctness}.
Moreover, we show that \modelname{} can be constructed even without the use of LLMs and updated incrementally as the corpus grows and evolves, making it practical and efficient for real deployment environments.
Together, these results suggest that improving how a corpus is organized, rather than only how agents search it, is a promising direction for agents operating over large and growing document collections.

%% file: Figures/fig_concept.tex
\begin{figure}[t!]
    \centering
    \includegraphics[width=\linewidth]{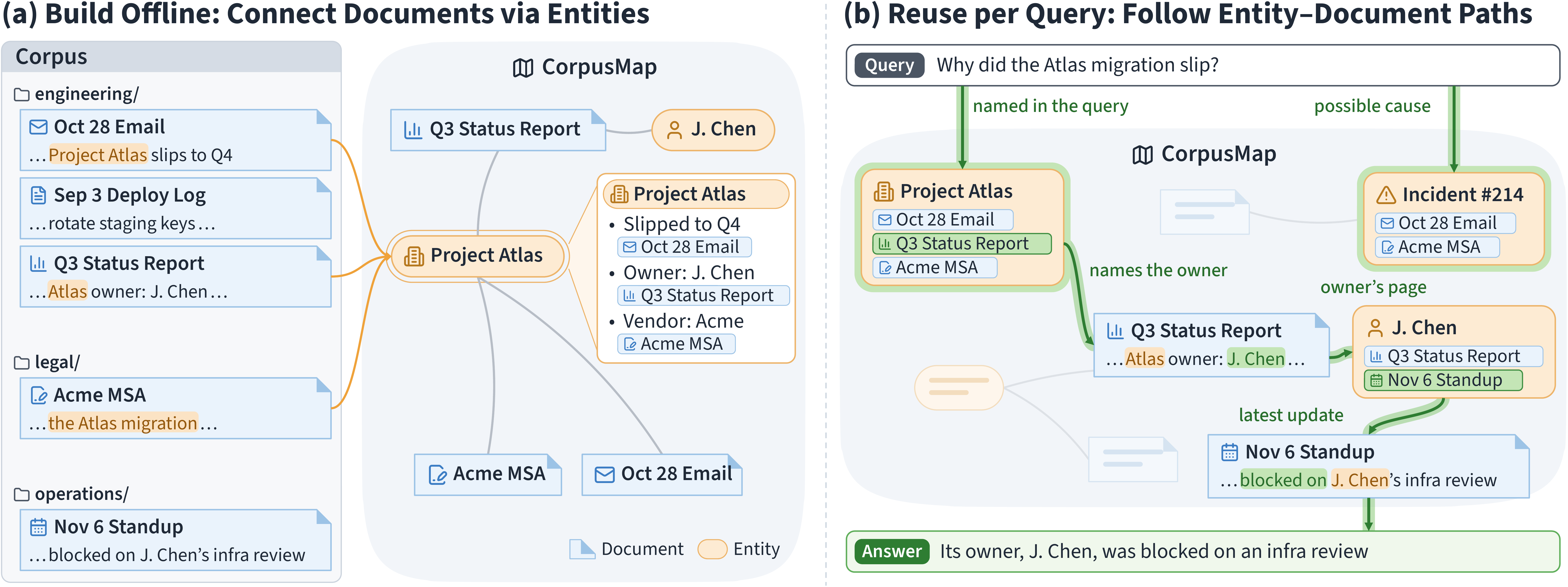}
    \vspace{-0.05in}
    \caption{Overview of \modelname{}.
    \textbf{(a)} Offline, recurring mentions of the same subject across documents in different folders are resolved into shared entities, which connect the documents into a reusable entity--document map. \textbf{(b)} For each query, the agent follows entity--document links in the same map from the entities relevant to the query to gather evidence for the answer. Green marks the walk; each entity card lists its source documents, highlighting the one opened next.
    }
    \label{fig:concept}
    \vspace{-0.15in}
\end{figure}

%% file: Sections/3_related_work.tex
\section{Related Work}
\label{sec:related_work}

\paragraph{Retrieval-Augmented Generation}
Retrieval-augmented generation (RAG) grounds language-model outputs in external knowledge by retrieving the documents or passages most relevant to a query, typically according to lexical or embedding similarity, and conditioning generation on the retrieved context~\citep{bm25, dpr, rag, rag_survey1}.
While simple RAG approaches score each reference independently, thereby discarding any inter-document information that might exist, graph-based approaches such as GraphRAG~\citep{graphrag} and HippoRAG~\citep{hipporag, hipporag2} organize the corpus into a graph of extracted entities and their relations, leveraging their shared context to retrieve content connected across documents.
Nevertheless, such structure is used within the retriever rather than exposed to the generator (the LLM), which typically still receives a fixed context selected in a single step (e.g., top-$k$ passages or graph-derived summaries), so that the generator must answer from whatever that retrieval returns and cannot reach a relevant document the retrieval misses, even when the answer depends on it.

\paragraph{Agentic Search and Corpus Interfaces for Agents}
To move beyond single-step retrieval, iterative and agentic approaches retrieve over multiple steps~\citep{ircot, flare, self-rag, adaptive-rag}, interleaving planning, search, source inspection, and tool-use throughout an LLM's reasoning trajectory~\citep{react, Search-o1, agentic-rag-survey1, dci, grepseek}.
Although these methods improve the query-time search policy, because the corpus itself remains a set of independent documents, any relationships inferred during one query response must be rediscovered for every question.
To address this, recent work organizes the corpus into agent-facing structures.
LLM-maintained wikis compile documents into cross-linked pages~\citep{llm-wiki, DBLP:journals/corr/abs-2605-25480}, but require the model to decide what becomes a page and how content is merged, decisions that can degrade as the corpus grows~\citep{Filesystem}.
Hierarchical skill trees organize documents into topical branches that an agent traverses to reach source documents~\citep{corpus2skill}, but assigning each document to only one or a few branches can separate evidence about the same subject across sources.

\paragraph{Entity Extraction, Linking, and Resolution}
Identifying entities in text has long been studied through named entity recognition~\citep{ner, ner2, ner3}, recently extended to open entity types by LLMs and lightweight generalist encoders~\citep{UniversalNER, GoLLIE, GLiNER}, and through entity linking, which grounds mentions in a reference knowledge base such as Wikipedia~\citep{DBLP:conf/emnlp/WuPJRZ20, DBLP:conf/iclr/CaoI0P21, DBLP:journals/semweb/SevgiliSAPB22}.
When no such knowledge base covers the entities of interest, cross-document coreference and entity resolution instead cluster the mentions that refer to the same entity across sources~\citep{DBLP:conf/lrec/CybulskaV14, DBLP:conf/acl/BarhomSEBRD19, DBLP:journals/pvldb/0001LSDT20, DBLP:journals/csur/PapadakisSTP20, DBLP:conf/acl/CattanESJD21}, with recent approaches ranging from LLM prompting~\citep{DBLP:journals/pvldb/NarayanCOR22, DBLP:conf/adbis/PeetersB23, DBLP:conf/edbt/PeetersSB25, DBLP:journals/pacmmod/FuT0MKG25} to lightweight zero-shot linkers~\citep{glinker}.
Our work builds on this line of research, leveraging these capabilities to organize a corpus around its resolved cross-document entities as navigational anchors for LLM agents.

%% file: Sections/4_method.tex
\section{Method}
\label{sec:method}

In this section, we first formalize agentic question answering over large document collections, and then present \modelname{}, an entity-centric navigation layer that exposes reusable cross-document evidence paths to the agent, together with the offline protocol that constructs it.

\subsection{Preliminaries}
\label{sec:method_preliminaries}

\paragraph{Task Formulation}
Let $\gD$ denote a corpus containing documents drawn from different sources, and let $q$ denote a question whose answer requires combining evidence distributed across multiple documents.
We write the agentic question-answering process as $(\hat{a}_q,\hat{\gD}_q)=\texttt{Agent}(q,\gD)$, where $\hat{a}_q$ is the generated answer and $\hat{\gD}_q\subseteq\gD$ is the selected supporting set.
We assess the resulting trajectory along three complementary axes: answer quality, requiring $\hat{a}_q$ to be correct and complete; retrieval quality, requiring $\hat{\gD}_q$ to cover the documents the question actually depends on; and efficiency, favoring limited context consumption.

\paragraph{Raw-Corpus Agentic Search}
We first consider an agent operating directly over the raw corpus, which is exposed as a flat collection of documents.
Given $q$, the agent uses standard shell commands (e.g., \texttt{find} and \texttt{grep}) to search over $\gD$, reads promising documents, and updates the selected supporting set $\hat{\gD}_q$ as evidence accumulates.
However, although this interface gives access to the full corpus, it encodes no relations between documents, so once a relevant document is found, locating related evidence requires further search.
Consequently, $\hat{\gD}_q$ may omit documents that searching for $q$ does not return, and the context consumed to work out how documents relate is spent again for each subsequent question, even when the same documents are involved.

\subsection{\modelname{}: An Entity-Centric Navigation Layer}
\label{sec:method_map}

To address this limitation, we introduce \modelname{}, which makes relations between documents explicit through a map $\gG$ of the corpus that is constructed offline from $\gD$ and shared across questions, so that the agent operates at inference as $\texttt{Agent}(q,\gD;\gG)$.
We build $\gG$ around the entities mentioned in documents (e.g., people, projects, or incidents), which can be identified in each document independently of any question, so that the map can be built in advance.

\paragraph{Map Representation}
Let $\gE$ denote the \emph{cross-document entities} that \modelname{} retains from $\gD$, that is, those linked to more than one document, and for each $e\in\gE$, let the document neighborhood $\gN(e)\subseteq\gD$ contain the documents in which a mention was resolved to $e$ during construction (\cref{sec:method_construction}).
We represent each retained entity as an entity node and each document as a document node, and write $\gL$ for the links between these two node types, so that the map is the bipartite graph
\begin{equation}
    \gG
    =
    (\gE\cup\gD,\gL),
    \qquad
    \gL
    =
    \{(e,d)\mid e\in\gE,\ d\in\gN(e)\},
    \label{eq:entity_document_map}
\end{equation}
where each link $(e,d)\in\gL$ connects an entity node to a document node.
Since a document is linked to each retained entity it mentions, two documents that share an entity are connected through it, and a document can belong to several neighborhoods at once.
Meanwhile, documents linked to no retained entity remain in $\gD$ as isolated nodes, accessible through raw-corpus search.

\paragraph{Source-Grounded Entity Pages}
Each entity node $e\in\gE$ is exposed to the agent as an \emph{Entity Page} that consolidates what the documents in $\gN(e)$ state about $e$: a brief overview, key facts each tagged with the document it comes from, the names under which $e$ appears, and links to every document in $\gN(e)$.
Since these facts may come from documents in different sources, a single page can bring together complementary evidence that the raw corpus keeps apart.

\subsection{Constructing \modelname{}}
\label{sec:method_construction}

We construct $\gG$ offline in the four stages summarized in \Cref{alg:entity_map_construction}.

\input{Algorithms/alg_entity_map}

\paragraph{Cataloging (Line 1)}
The \emph{entity types} worth extracting, such as products, incidents, or configuration flags, vary from one corpus to another and cannot be exhaustively specified in advance.
We therefore induce a catalog $\gC$ of these types from the corpus itself with an LLM, proposing a candidate catalog from each of several small sets of sampled documents, synthesizing these candidates into one, and verifying and revising the result; the resulting catalog, which specifies a name, definition, identity criteria, and observed examples for each type, is then fixed for the remaining stages.

\paragraph{Extraction (Line 2)}
Since the catalog specifies the kinds of entities in the map (e.g., \emph{Project}) but not the instances present in the corpus (e.g., ``Project Atlas''), we use it together with the surrounding document context to identify and type \emph{entity mentions}, grouping those denoting the same subject within a document into a single document-local entity.

\paragraph{Resolution (Lines 3--13)}
We ground every extracted name in its source-text occurrence before resolving the local entities of each document, in turn, against a shared registry that starts out empty: for each of them, we retrieve plausible registry entries and weigh the current document against the candidate evidence to either \texttt{LINK} the observation to an existing entry, \texttt{ADD} a new entity, or leave it \texttt{UNRESOLVED}.
Each \texttt{LINK} or \texttt{ADD} records a grounded entity--document link, whereas \texttt{UNRESOLVED} observations create none.

\paragraph{Rendering (Lines 14--21)}
We keep only the cross-document entities, that is, those linked to at least two documents, since only these provide reusable navigational paths, and render the neighborhood $\gN(e)$ of each retained entity as its Entity Page.
With the source documents and the links between them, these pages yield the map $\gG$ of \cref{eq:entity_document_map}.
The Extraction, Resolution, and Rendering stages can be instantiated with an LLM or with off-the-shelf and deterministic alternatives.

\subsection{Navigating \modelname{}}
\label{sec:method_navigation}

We now describe how the agent accesses the map at inference.
Specifically, each Entity Page is stored as a file alongside the raw documents and lists the file paths of its linked documents, so that the agent can read and search the map with the same tools as the raw corpus.
Along with the question, the agent receives, as candidates, the file paths of the documents linked to the Entity Pages relevant to the question, and decides which of them to read.
The agent can further search both Entity Pages and documents, following a link from a page by reading a listed file, or from a document by searching for the pages that list it (\cref{fig:concept}).

%% file: Algorithms/alg_entity_map.tex
\begin{algorithm}[t]
\caption{\modelname{} construction protocol.}
\label{alg:entity_map_construction}
\small
\begin{algorithmic}[1]
\Require Document corpus $\gD$, entity-processing backend, page renderer
\Ensure Entity-centric map $\gG$ with Entity Pages
\State $\gC \gets \Call{InduceTypeCatalog}{\gD}$ \Comment{propose from sampled documents, synthesize, verify, revise}
\State $\{\gE_d^{\mathrm{local}}\}_{d\in\gD} \gets \Call{ExtractLocalEntities}{\gD,\gC}$ \Comment{identify, type, and group mentions per document}
\State $\gR \gets \emptyset,\ \gL \gets \emptyset$ \Comment{empty registry $\gR$ and link set $\gL$}
\For{each document $d\in\gD$ and each local entity $z\in\gE_d^{\mathrm{local}}$}
    \State $\delta \gets \Call{Resolve}{z,\gR}$ \Comment{retrieve candidates from $\gR$ and decide}
    \If{$\delta=\mathtt{LINK}(e)$}
        \State $\Call{UpdateEntity}{\gR,e,z}$; $\gL \gets \gL\cup\{(e,d)\}$
    \ElsIf{$\delta=\mathtt{ADD}$}
        \State $e \gets \Call{AddEntity}{\gR,z}$; $\gL \gets \gL\cup\{(e,d)\}$
    \Else
        \State $\Call{RecordUnresolved}{z,d}$
    \EndIf
\EndFor
\State $\gE \gets \Call{CrossDocumentEntities}{\gR,\gL}$ \Comment{linked to at least two documents}
\State $\gL \gets \{(e,d)\in\gL\mid e\in\gE\}$
\For{each entity $e\in\gE$}
    \State $\gN(e) \gets \{d\in\gD\mid(e,d)\in\gL\}$
    \State $p_e \gets \Call{RenderEntityPage}{e,\gN(e)}$ \Comment{facts about $e$ grounded in and linked to $\gN(e)$}
\EndFor
\State $\gG \gets (\gE\cup\gD,\gL)$
\State \Return $\gG$ with $\{p_e\}_{e\in\gE}$
\end{algorithmic}
\end{algorithm}

%% file: Sections/5_experimental_setup.tex
\section{Experimental Setup}
\label{sec:experimental_setup}

We now describe the benchmarks and evaluation, baselines, and implementation details.

\paragraph{Benchmarks and Evaluation}
To evaluate \modelname{}, we use three benchmarks that contain questions requiring evidence distributed across multiple documents: \textbf{EnterpriseRAG-Bench}~\citep{enterpriseragbench}, \textbf{WixQA}~\citep{WixQA}, and \textbf{HERB}~\citep{HERB}.
Specifically, we use the 80 questions in the categories of EnterpriseRAG-Bench that consist entirely of multi-document questions, the 79 multi-document questions of WixQA, and the 238 content-based questions of HERB, which ask about information stated across multiple documents.
For the corpora, we use fixed sets of 2,819 and 6,365 documents for EnterpriseRAG-Bench and HERB, respectively, both including all gold documents, and the full set of 6,221 articles for WixQA.
Following the three axes in \cref{sec:method_preliminaries}, we evaluate (1) \textbf{answer quality}: correctness, completeness, factuality, and content; (2) \textbf{retrieval quality}: document recall and context recall; and (3) \textbf{efficiency}: the input tokens accumulated over the full agent trajectory per question.

\paragraph{Baselines and Our Method}
We compare \modelname{} against \textsc{Raw Corpus} and four baselines that organize the same corpus around different units, while the original documents remain accessible to the agent in every method.
\textbf{\textsc{Raw Corpus}} adds no navigation layer to the documents.
\textbf{\textsc{Document Page}} represents each document by an LLM-generated page of its key facts, \textbf{\textsc{Group Page}} consolidates the documents within each group defined by the corpus itself (e.g., its folders) into a single page, and \textbf{\textsc{LLM Wiki}}~\citep{llm-wiki} lets the LLM freely write cross-linked pages over the corpus.
\textbf{\textsc{Corpus2Skill}}~\citep{corpus2skill} organizes the corpus into a topical hierarchy of LLM-summarized document clusters.
\textbf{\modelname{} (Ours)} organizes the corpus into an entity-centric map of Entity Pages linked to their source documents.
We also report \textbf{\textsc{Gold Documents (Oracle)}}, which provides only the gold documents to the model as reference for the model capability ceiling under idealized retrieval.

\paragraph{Implementation Details}
For the main results, we use four GPT models spanning a wide range of costs, GPT-5.5~\citep{gpt55-system-card} and GPT-5.6 Luna, Terra, and Sol~\citep{gpt56-system-card}, where the same LLM constructs the artifacts of each method and serves as the agent answering the questions.
For the analyses beyond the main results, we mainly use the most and least expensive of them, GPT-5.5 and GPT-5.6 Luna.
We additionally use DeepSeek-V4-Pro~\citep{DeepSeek-V4} and MAI-Thinking-1~\citep{mai-thinking-1}, as well as the open-weight Qwen3.8-27B~\citep{qwen38}.
For LLM-judged metrics, we use GPT-5.6 Sol.
Following \citet{enterpriseragbench}, we use a terminal-based agent that navigates the corpus through shell commands, under their per-question execution budget.
Please refer to \cref{sec:appendix_experiment} for more details.

%% file: Sections/6_experimental_results.tex
\section{Experimental Results and Analyses}
\label{sec:experimental_results}

We first examine the effectiveness of \modelname{}, and then analyze its practical aspects, with further analyses provided in \cref{sec:appendix_results}.

\subsection{Effectiveness of \modelname{}}
\label{sec:results_effectiveness}

\input{Tables/tab_main}

\paragraph{Main Results}
\Cref{tab:main_results} presents the main results, showing that \modelname{} consistently achieves the best answer and retrieval quality across all benchmarks and LLMs, with significant overall gains over every baseline (\cref{tab:significance}), at a lower average cost per query than raw-corpus agentic search (\cref{fig:enterprise_rag_correctness}).
Notably, the four baselines that organize the corpus in other ways do not consistently improve over \textsc{Raw Corpus}, indicating that simply adding a navigation layer does not guarantee improvement.
Moreover, \modelname{} improves quality while reducing tokens, with the largest savings for GPT-5.5 and GPT-5.6 Sol, the two most expensive LLMs.
Also, \modelname{} substantially narrows the gap to the non-comparable Oracle, reflecting its effectiveness in gathering cross-document evidence.
Finally, \modelname{} also outperforms \textsc{Raw Corpus} with fewer tokens on the remaining questions of EnterpriseRAG-Bench, most of which are grounded in a single document (\cref{tab:enterprise_remaining}), indicating that the map remains beneficial even when cross-document evidence is not required.

\input{Tables/tab_main_deepseek_mai_common_reader}

\paragraph{Generalization to Other Model Families}
To examine whether the effectiveness of \modelname{} generalizes beyond the GPT family, we further evaluate it with DeepSeek-V4-Pro and MAI-Thinking-1, and report the results in \cref{tab:main_results_deepseek_mai}.
We find that \modelname{} again achieves the best Overall Quality with both LLMs, whereas the baselines that organize the corpus in other ways do not consistently improve over \textsc{Raw Corpus}, as observed with the GPT models.
This indicates that the benefit of exposing cross-document connections through the entity-centric map is not tied to a particular model family, but carries over to LLMs of different architectures.

\paragraph{Comparison with Retrieval-Based Approaches}
We also compare against BM25~\citep{bm25}, dense retrieval~\citep{text-embedding-3}, HippoRAG~\citep{hipporag, hipporag2}, and GraphRAG~\citep{graphrag} under the retrieve-then-generate paradigm, where the LLM answers directly from a fixed amount of context retrieved for the question, without accessing the full corpus itself.
As shown in \cref{tab:common_reader_results}, \modelname{} outperforms all of them with both LLMs, including the graph-based HippoRAG and GraphRAG.
This suggests that answering from a retrieved context alone is limited by what the retrieval returns, whereas exposing cross-document connections to the agent lets it continue to gather the missing evidence.

\paragraph{Case Study}
We present a case study example in \cref{tab:case_study_streaming}.
Given the question of how signing is represented in the v1 specification of a manifest, the answer lies in two documents stored in different sources: an earlier draft and the v1 specification that revised it.
\textsc{Corpus2Skill}, which organizes the corpus into a tree structure, places both documents under a single cluster, and its agent explores several other branches but fails to locate them, concluding that no such document exists.
In contrast, \modelname{} links each document to several Entity Pages, and its agent navigates between Entity Pages and documents: (1) opening the page of a work item that links the draft, (2) reading the draft and searching the Entity Pages with its terms, and (3) opening the \texttt{Serving Runtime} page that links the v1 specification and reading it.
This path enables the agent to answer with the fields defined in v1, highlighting how linking documents through the entities they share offers multiple paths to the same evidence, whereas a tree structure places each document mainly under a single branch.

\subsection{Practical Aspects of \modelname{}}
\label{sec:results_practical}

We first examine how the map, once constructed, is reused across queries, LLMs, and corpus updates, and then whether it remains effective when constructed with off-the-shelf tools, over larger corpora, and with open-weight models.

\paragraph{Amortized Construction Cost}
In addition to the cost per query, \modelname{} requires a one-time cost to construct the map, which is then shared by all subsequent queries over the same corpus.
To see how this cost is amortized, we add the construction cost, divided by the number of queries the map serves, to the cost per query of \modelname{}.
As shown in \cref{fig:amortized_cost}, although the amortized cost of \modelname{} is initially higher than that of \textsc{Raw Corpus}, it decreases as more queries are served and becomes lower beyond a certain number of queries for every LLM.
This suggests that the construction cost of the map is recovered as it is reused across queries.

\input{Tables/tab_map_transfer}
\input{Figures/fig_evolving_corpora}

\paragraph{Map Reuse Across LLMs}
We further examine whether a map constructed by one LLM can be reused by another LLM at inference time, and report the results in \cref{tab:map_transfer}.
We find that even the map constructed by the least expensive LLM, at a small fraction of the cost of the most expensive one, improves the overall quality over \textsc{Raw Corpus} for every answering LLM.
This suggests that the map remains useful beyond the LLM that constructs it, even across model families, allowing it to be built once with an inexpensive LLM and reused by stronger ones.

\paragraph{Incremental Map Updates}
To examine whether \modelname{} can be maintained as new documents arrive, we order the EnterpriseRAG-Bench corpus chronologically by the timestamps provided in the benchmark and incrementally extend the map over its growing prefix, reusing the existing registry and re-rendering only the Entity Pages whose linked documents change, instead of rebuilding the map from scratch, while the agent can search the full raw corpus at every state.
As shown in \cref{fig:evolving_corpora}(a), each incremental update saves a substantial fraction of the construction tokens of a full rebuild, and \cref{fig:evolving_corpora}(b) shows that quality improves overall with each update, with the final map performing comparably to a full rebuild over the same documents.
Also, \cref{fig:evolving_corpora}(c) shows that the questions whose evidence is incorporated by an update improve after it, even though they were already searchable in the raw corpus, indicating that incorporating new documents into the map matters beyond making them accessible.

\input{Figures/fig_corpus_scale}

\paragraph{Off-the-Shelf Entity Construction}
We instantiate the Extraction, Resolution, and Rendering stages of \modelname{} with off-the-shelf and deterministic components, where GLinker~\citep{glinker} extracts and links entity mentions with open GLiNER models and Entity Pages are rendered from the extracted evidence without any LLM calls.
As shown in \cref{fig:corpus_scale}, the resulting map is comparably effective to the LLM-constructed map, and outperforms \textsc{Raw Corpus} on every metric at a lower cost, indicating that \modelname{} can be instantiated with off-the-shelf tools as well as with LLMs.

\paragraph{Scaling Corpora}
We further examine \modelname{} as the corpus grows, expanding the EnterpriseRAG-Bench corpus with distractor documents while keeping all gold documents.
As shown in \cref{fig:corpus_scale}, \modelname{} outperforms \textsc{Raw Corpus} at every corpus size while using fewer tokens, so that its advantage persists at scale.

\paragraph{Open-Weight Models}
We further evaluate \modelname{} with the open-weight Qwen3.8-27B over the same map constructed with GLinker.
As shown in \cref{fig:corpus_scale}, \modelname{} improves over \textsc{Raw Corpus} with Qwen on every metric and corpus size while using fewer tokens, indicating that its benefits are not limited to proprietary LLMs.

%% file: Tables/tab_main.tex
\begin{table*}[t!]
\caption{Main results across EnterpriseRAG-Bench, WixQA, and HERB with GPT-5.5 and GPT-5.6 models, as means $\pm$ standard deviations over three runs. Overall reports dataset-balanced quality and geometric-mean input-token ratios to Raw Corpus within each LLM. Best and second-best effectiveness scores per LLM among retrieval methods are \textbf{bolded} and \underline{underlined}.}
\label{tab:main_results}
\vspace{-0.05in}
\centering
\small
\ADLactivate
\renewcommand{\arraystretch}{0.98}
\setlength{\tabcolsep}{1.0pt}
\scalebox{0.677}{%
\begin{tabular}{c l c c c c c c c c c >{\columncolor{gray!10}}c >{\columncolor{gray!10}}c}
\toprule
& & \multicolumn{4}{c}{\textbf{EnterpriseRAG-Bench}}
& \multicolumn{3}{c}{\textbf{WixQA}}
& \multicolumn{2}{c}{\textbf{HERB}} & \multicolumn{2}{c}{\textbf{Overall}} \\

\cmidrule(lr){3-6} \cmidrule(lr){7-9} \cmidrule(lr){10-11} \cmidrule(lr){12-13}
& \textbf{Method}
& \textbf{Doc. Rec.} $\uparrow$
& \textbf{Correct.} $\uparrow$
& \textbf{Complete.} $\uparrow$
& \textbf{Tokens} $\downarrow$
& \textbf{Ctx. Rec.} $\uparrow$ & \textbf{Fact.} $\uparrow$ & \textbf{Tokens} $\downarrow$ & \textbf{Content} $\uparrow$ & \textbf{Tokens} $\downarrow$ & \multicolumn{1}{c}{\textbf{Quality} $\uparrow$} & \multicolumn{1}{c}{\textbf{Rel. Tok.} $\downarrow$} \\
\midrule
\midrule
\multirow{7}{*}{\rotatebox{90}{\textbf{GPT-5.5}}}
& \textsc{Raw Corpus}
& 61.62 {\textsmaller{$\pm$ 1.66}}
& \underline{62.08} {\textsmaller{$\pm$ 4.12}}
& \underline{73.11} {\textsmaller{$\pm$ 0.98}}
& 206.5k
& 73.31 {\textsmaller{$\pm$ 0.75}} & \underline{67.51} {\textsmaller{$\pm$ 0.39}} & 337.2k & 62.31 {\textsmaller{$\pm$ 0.59}} & 598.4k & 66.11 & 1.00$\times$ \\
& \textsc{Document Page}
& 64.18 {\textsmaller{$\pm$ 1.55}}
& 57.50 {\textsmaller{$\pm$ 1.02}}
& 68.97 {\textsmaller{$\pm$ 1.83}}
& 175.6k
& \underline{77.85} {\textsmaller{$\pm$ 1.18}} & 66.98 {\textsmaller{$\pm$ 1.90}} & 288.0k & \underline{63.26} {\textsmaller{$\pm$ 0.14}} & 680.9k & \underline{66.41} & 0.94$\times$ \\
& \textsc{Group Page}
& \underline{64.75} {\textsmaller{$\pm$ 0.96}}
& 48.75 {\textsmaller{$\pm$ 2.70}}
& 62.62 {\textsmaller{$\pm$ 2.58}}
& 423.0k
& 60.86 {\textsmaller{$\pm$ 3.68}} & 62.24 {\textsmaller{$\pm$ 4.45}} & 1,125.8k & 62.97 {\textsmaller{$\pm$ 1.66}} & 392.9k & 61.07 & 1.65$\times$ \\
& \textsc{LLM Wiki}
& 63.15 {\textsmaller{$\pm$ 1.55}}
& 56.67 {\textsmaller{$\pm$ 2.36}}
& 67.74 {\textsmaller{$\pm$ 1.77}}
& 192.4k
& 73.73 {\textsmaller{$\pm$ 0.52}} & 65.19 {\textsmaller{$\pm$ 0.26}} & 140.9k & 61.49 {\textsmaller{$\pm$ 0.76}} & 387.0k & 64.49 & 0.63$\times$ \\
& \textsc{Corpus2Skill}
& 49.39 {\textsmaller{$\pm$ 2.32}}
& 47.50 {\textsmaller{$\pm$ 2.70}}
& 57.29 {\textsmaller{$\pm$ 1.01}}
& 88.8k
& 65.93 {\textsmaller{$\pm$ 2.07}} & 60.34 {\textsmaller{$\pm$ 0.39}} & 76.0k & 21.94 {\textsmaller{$\pm$ 0.79}} & 180.9k & 45.49 & 0.31$\times$ \\
\rowcolor{oursrow}\cellcolor{white}
& \textbf{\modelname{} (Ours)}
& \textbf{76.17 {\textsmaller{$\pm$ 0.30}}}
& \textbf{73.75 {\textsmaller{$\pm$ 1.77}}}
& \textbf{79.88 {\textsmaller{$\pm$ 1.00}}}
& 88.1k
& \textbf{82.70 {\textsmaller{$\pm$ 0.79}}} & \textbf{70.68 {\textsmaller{$\pm$ 0.83}}} & 74.5k & \textbf{64.37 {\textsmaller{$\pm$ 0.89}}} & 494.7k & \textbf{72.55} & 0.43$\times$ \\
\noalign{\vskip 0.25ex}\cdashline{2-13}\noalign{\vskip 0.75ex}
& \textit{Gold Documents (Oracle)}
& \textit{100.00 {\textsmaller{$\pm$ 0.00}}}
& \textit{87.08 {\textsmaller{$\pm$ 2.12}}}
& \textit{81.01 {\textsmaller{$\pm$ 0.59}}}
& \textit{7.1k}
& \textit{85.97 {\textsmaller{$\pm$ 0.30}}} & \textit{79.22 {\textsmaller{$\pm$ 0.30}}} & \textit{2.1k} & \textit{66.90 {\textsmaller{$\pm$ 0.84}}} & \textit{11.4k} & \textit{79.62} & \textit{0.02$\times$} \\
\midrule
\multirow{7}{*}{\rotatebox{90}{\textbf{Luna}}}
& \textsc{Raw Corpus}
& 56.40 {\textsmaller{$\pm$ 0.79}}
& \underline{60.42} {\textsmaller{$\pm$ 5.14}}
& \underline{73.12} {\textsmaller{$\pm$ 0.96}}
& 80.7k
& 47.36 {\textsmaller{$\pm$ 1.66}} & 59.39 {\textsmaller{$\pm$ 1.04}} & 129.3k & 44.86 {\textsmaller{$\pm$ 0.37}} & 219.4k & 53.85 & 1.00$\times$ \\
& \textsc{Document Page}
& \underline{56.96} {\textsmaller{$\pm$ 2.11}}
& 50.42 {\textsmaller{$\pm$ 3.28}}
& 68.78 {\textsmaller{$\pm$ 2.40}}
& 160.0k
& \underline{61.92} {\textsmaller{$\pm$ 0.91}} & \underline{61.18} {\textsmaller{$\pm$ 2.23}} & 88.7k & 46.91 {\textsmaller{$\pm$ 0.49}} & 222.4k & \underline{55.73} & 1.11$\times$ \\
& \textsc{Group Page}
& 53.95 {\textsmaller{$\pm$ 1.82}}
& 42.08 {\textsmaller{$\pm$ 1.18}}
& 58.82 {\textsmaller{$\pm$ 1.30}}
& 347.6k
& 43.57 {\textsmaller{$\pm$ 0.79}} & 55.59 {\textsmaller{$\pm$ 2.71}} & 145.3k & 41.47 {\textsmaller{$\pm$ 0.24}} & 119.1k & 47.55 & 1.38$\times$ \\
& \textsc{LLM Wiki}
& 54.54 {\textsmaller{$\pm$ 3.72}}
& 49.17 {\textsmaller{$\pm$ 2.12}}
& 65.90 {\textsmaller{$\pm$ 1.68}}
& 144.5k
& 58.12 {\textsmaller{$\pm$ 2.15}} & 59.70 {\textsmaller{$\pm$ 2.07}} & 117.3k & \underline{47.71} {\textsmaller{$\pm$ 0.55}} & 213.3k & 54.39 & 1.16$\times$ \\
& \textsc{Corpus2Skill}
& 35.73 {\textsmaller{$\pm$ 1.25}}
& 31.25 {\textsmaller{$\pm$ 1.77}}
& 49.90 {\textsmaller{$\pm$ 0.20}}
& 102.4k
& 56.33 {\textsmaller{$\pm$ 3.66}} & 53.80 {\textsmaller{$\pm$ 3.05}} & 56.2k & 17.78 {\textsmaller{$\pm$ 0.55}} & 145.7k & 37.27 & 0.72$\times$ \\
\rowcolor{oursrow}\cellcolor{white}
& \textbf{\modelname{} (Ours)}
& \textbf{75.11 {\textsmaller{$\pm$ 0.27}}}
& \textbf{67.08 {\textsmaller{$\pm$ 4.25}}}
& \textbf{77.42 {\textsmaller{$\pm$ 1.45}}}
& 60.1k
& \textbf{76.90 {\textsmaller{$\pm$ 0.93}}} & \textbf{69.30 {\textsmaller{$\pm$ 0.26}}} & 41.0k & \textbf{50.45 {\textsmaller{$\pm$ 0.27}}} & 254.2k & \textbf{65.58} & 0.65$\times$ \\
\noalign{\vskip 0.25ex}\cdashline{2-13}\noalign{\vskip 0.75ex}
& \textit{Gold Documents (Oracle)}
& \textit{100.00 {\textsmaller{$\pm$ 0.00}}}
& \textit{82.50 {\textsmaller{$\pm$ 4.68}}}
& \textit{76.90 {\textsmaller{$\pm$ 0.57}}}
& \textit{7.1k}
& \textit{85.76 {\textsmaller{$\pm$ 0.52}}} & \textit{75.53 {\textsmaller{$\pm$ 0.60}}} & \textit{2.1k} & \textit{72.90 {\textsmaller{$\pm$ 0.66}}} & \textit{11.4k} & \textit{80.00} & \textit{0.04$\times$} \\
\midrule
\multirow{7}{*}{\rotatebox{90}{\textbf{Terra}}}
& \textsc{Raw Corpus}
& \underline{67.13} {\textsmaller{$\pm$ 1.13}}
& \underline{64.58} {\textsmaller{$\pm$ 2.36}}
& \underline{77.11} {\textsmaller{$\pm$ 0.81}}
& 70.9k
& 65.08 {\textsmaller{$\pm$ 1.42}} & 62.45 {\textsmaller{$\pm$ 0.91}} & 354.6k & \underline{63.88} {\textsmaller{$\pm$ 0.74}} & 288.3k & \underline{65.75} & 1.00$\times$ \\
& \textsc{Document Page}
& 62.16 {\textsmaller{$\pm$ 3.32}}
& 56.67 {\textsmaller{$\pm$ 4.12}}
& 69.39 {\textsmaller{$\pm$ 3.41}}
& 191.7k
& \underline{72.78} {\textsmaller{$\pm$ 2.37}} & \underline{64.77} {\textsmaller{$\pm$ 1.47}} & 138.9k & 57.22 {\textsmaller{$\pm$ 2.15}} & 274.9k & 62.91 & 1.00$\times$ \\
& \textsc{Group Page}
& 62.00 {\textsmaller{$\pm$ 1.69}}
& 38.33 {\textsmaller{$\pm$ 2.95}}
& 52.57 {\textsmaller{$\pm$ 1.10}}
& 355.3k
& 26.69 {\textsmaller{$\pm$ 6.88}} & 43.14 {\textsmaller{$\pm$ 5.67}} & 324.6k & 45.62 {\textsmaller{$\pm$ 0.49}} & 150.7k & 43.84 & 1.34$\times$ \\
& \textsc{LLM Wiki}
& 58.77 {\textsmaller{$\pm$ 0.66}}
& 43.33 {\textsmaller{$\pm$ 2.57}}
& 64.58 {\textsmaller{$\pm$ 0.31}}
& 161.6k
& 66.03 {\textsmaller{$\pm$ 2.10}} & 61.60 {\textsmaller{$\pm$ 3.03}} & 129.5k & 56.77 {\textsmaller{$\pm$ 1.41}} & 390.7k & 58.71 & 1.04$\times$ \\
& \textsc{Corpus2Skill}
& 45.84 {\textsmaller{$\pm$ 0.43}}
& 44.58 {\textsmaller{$\pm$ 2.57}}
& 55.97 {\textsmaller{$\pm$ 0.38}}
& 153.7k
& 67.30 {\textsmaller{$\pm$ 1.94}} & 57.81 {\textsmaller{$\pm$ 1.94}} & 72.4k & 20.23 {\textsmaller{$\pm$ 0.43}} & 185.8k & 43.86 & 0.66$\times$ \\
\rowcolor{oursrow}\cellcolor{white}
& \textbf{\modelname{} (Ours)}
& \textbf{73.08 {\textsmaller{$\pm$ 1.98}}}
& \textbf{73.75 {\textsmaller{$\pm$ 1.77}}}
& \textbf{79.83 {\textsmaller{$\pm$ 0.92}}}
& 76.2k
& \textbf{80.91 {\textsmaller{$\pm$ 1.22}}} & \textbf{71.73 {\textsmaller{$\pm$ 0.98}}} & 96.2k & \textbf{65.65 {\textsmaller{$\pm$ 0.70}}} & 289.4k & \textbf{72.51} & 0.66$\times$ \\
\noalign{\vskip 0.25ex}\cdashline{2-13}\noalign{\vskip 0.75ex}
& \textit{Gold Documents (Oracle)}
& \textit{100.00 {\textsmaller{$\pm$ 0.00}}}
& \textit{82.08 {\textsmaller{$\pm$ 0.59}}}
& \textit{81.17 {\textsmaller{$\pm$ 0.36}}}
& \textit{7.1k}
& \textit{85.55 {\textsmaller{$\pm$ 0.39}}} & \textit{77.64 {\textsmaller{$\pm$ 0.98}}} & \textit{2.1k} & \textit{70.37 {\textsmaller{$\pm$ 0.83}}} & \textit{11.4k} & \textit{79.90} & \textit{0.03$\times$} \\
\midrule
\multirow{7}{*}{\rotatebox{90}{\textbf{Sol}}}
& \textsc{Raw Corpus}
& 64.79 {\textsmaller{$\pm$ 2.43}}
& 67.08 {\textsmaller{$\pm$ 3.28}}
& \underline{76.75} {\textsmaller{$\pm$ 1.85}}
& 55.5k
& 69.30 {\textsmaller{$\pm$ 1.34}} & 68.35 {\textsmaller{$\pm$ 1.81}} & 282.3k & \underline{64.69} {\textsmaller{$\pm$ 1.42}} & 430.2k & \underline{67.69} & 1.00$\times$ \\
& \textsc{Document Page}
& \underline{67.64} {\textsmaller{$\pm$ 0.94}}
& \underline{67.50} {\textsmaller{$\pm$ 2.04}}
& 72.13 {\textsmaller{$\pm$ 1.52}}
& 55.2k
& \underline{73.84} {\textsmaller{$\pm$ 1.33}} & \underline{70.78} {\textsmaller{$\pm$ 1.76}} & 1,496.8k & 57.53 {\textsmaller{$\pm$ 3.54}} & 5,003.6k & 66.31 & 3.94$\times$ \\
& \textsc{Group Page}
& 59.31 {\textsmaller{$\pm$ 2.44}}
& 43.75 {\textsmaller{$\pm$ 2.70}}
& 56.91 {\textsmaller{$\pm$ 3.42}}
& 188.5k
& 18.67 {\textsmaller{$\pm$ 0.26}} & 39.77 {\textsmaller{$\pm$ 1.94}} & 458.6k & 11.14 {\textsmaller{$\pm$ 0.52}} & 59.0k & 31.23 & 0.91$\times$ \\
& \textsc{LLM Wiki}
& 60.70 {\textsmaller{$\pm$ 1.08}}
& 59.17 {\textsmaller{$\pm$ 4.12}}
& 69.83 {\textsmaller{$\pm$ 0.42}}
& 53.2k
& 70.46 {\textsmaller{$\pm$ 1.04}} & 68.14 {\textsmaller{$\pm$ 0.54}} & 68.7k & 62.91 {\textsmaller{$\pm$ 1.74}} & 208.5k & 65.15 & 0.48$\times$ \\
& \textsc{Corpus2Skill}
& 45.10 {\textsmaller{$\pm$ 0.72}}
& 42.08 {\textsmaller{$\pm$ 1.18}}
& 53.78 {\textsmaller{$\pm$ 1.89}}
& 126.4k
& 64.66 {\textsmaller{$\pm$ 1.42}} & 59.49 {\textsmaller{$\pm$ 1.61}} & 99.1k & 24.24 {\textsmaller{$\pm$ 0.70}} & 159.0k & 44.44 & 0.67$\times$ \\
\rowcolor{oursrow}\cellcolor{white}
& \textbf{\modelname{} (Ours)}
& \textbf{71.46 {\textsmaller{$\pm$ 0.41}}}
& \textbf{80.83 {\textsmaller{$\pm$ 0.59}}}
& \textbf{81.80 {\textsmaller{$\pm$ 0.60}}}
& 56.2k
& \textbf{83.33 {\textsmaller{$\pm$ 0.83}}} & \textbf{72.78 {\textsmaller{$\pm$ 1.18}}} & 52.0k & \textbf{67.97 {\textsmaller{$\pm$ 1.08}}} & 186.3k & \textbf{74.68} & 0.43$\times$ \\
\noalign{\vskip 0.25ex}\cdashline{2-13}\noalign{\vskip 0.75ex}
& \textit{Gold Documents (Oracle)}
& \textit{100.00 {\textsmaller{$\pm$ 0.00}}}
& \textit{80.42 {\textsmaller{$\pm$ 1.18}}}
& \textit{80.85 {\textsmaller{$\pm$ 0.54}}}
& \textit{7.1k}
& \textit{85.97 {\textsmaller{$\pm$ 0.65}}} & \textit{79.85 {\textsmaller{$\pm$ 0.30}}} & \textit{2.1k} & \textit{69.08 {\textsmaller{$\pm$ 0.77}}} & \textit{11.4k} & \textit{79.69} & \textit{0.03$\times$} \\
\bottomrule
\end{tabular}%
}
\vspace{-0.1in}
\end{table*}

%% file: Tables/tab_main_deepseek_mai_common_reader.tex
\begin{table}[t!]
    \begin{minipage}[t]{0.575\linewidth}
        \input{Tables/tab_main_deepseek_mai}
    \end{minipage}
    \hfill
    \begin{minipage}[t]{0.395\linewidth}
        \input{Tables/tab_common_reader}
    \end{minipage}
\vspace{-0.1in}
\end{table}

%% file: Tables/tab_main_deepseek_mai.tex
\caption{Overall Quality with LLMs from other model families on EnterpriseRAG-Bench.}
\label{tab:main_results_deepseek_mai}
\vspace{-0.05in}
\centering
\small
\renewcommand{\arraystretch}{0.98}
\setlength{\tabcolsep}{3pt}
\begin{tabular}{l c c c c}
\toprule
& \multicolumn{2}{c}{\textbf{DeepSeek}} & \multicolumn{2}{c}{\textbf{MAI}} \\
\cmidrule(lr){2-3} \cmidrule(lr){4-5}
\textbf{Method} & \textbf{Quality} & \textbf{Tokens} & \textbf{Quality} & \textbf{Tokens} \\
\midrule
\midrule
\textsc{Raw Corpus} & \underline{68.02} & 1,349.4k & 34.06 & 129.6k \\
\textsc{Document Page} & 43.18 & 4,047.0k & \underline{38.56} & 208.4k \\
\textsc{Group Page} & 59.24 & 1,965.5k & 32.81 & 421.7k \\
\textsc{LLM Wiki} & 57.99 & 3,523.8k & 35.57 & 109.8k \\
\textsc{Corpus2Skill} & 49.72 & 628.7k & 21.33 & 120.4k \\
\midrule
\textbf{\modelname{} (Ours)} & \textbf{71.11} & 1,043.3k & \textbf{45.07} & 158.8k \\
\bottomrule
\end{tabular}

%% file: Tables/tab_common_reader.tex
\caption{Retrieval-based approaches on EnterpriseRAG-Bench.}
\label{tab:common_reader_results}
\vspace{-0.05in}
\centering
\small
\renewcommand{\arraystretch}{0.98}
\setlength{\tabcolsep}{3pt}
\begin{tabular}{l c c}
\toprule
& \multicolumn{2}{c}{\textbf{Quality}} \\
\cmidrule(lr){2-3}
\textbf{Method} & \textbf{GPT-5.5} & \textbf{Luna} \\
\midrule
\midrule
\textsc{BM25} & 63.66 & 60.78 \\
\textsc{Dense} & 65.05 & 58.84 \\
\textsc{HippoRAG} & 64.66 & 61.74 \\
\textsc{GraphRAG} & 47.95 & 43.46 \\
\midrule
\textsc{Raw Corpus} & \underline{65.60} & \underline{63.31} \\
\textbf{\modelname{} (Ours)} & \textbf{76.60} & \textbf{73.20} \\
\bottomrule
\end{tabular}

%% file: Tables/tab_map_transfer.tex
\begin{table}[t!]
\caption{Overall Quality of map reuse across LLMs on EnterpriseRAG-Bench. Rows denote the map builder, columns the answering LLM, and Cost the one-time construction cost.}
\label{tab:map_transfer}
\vspace{-0.05in}
\centering
\small
\renewcommand{\arraystretch}{0.98}
\setlength{\tabcolsep}{15.5pt}
\begin{tabular}{l c c c c r}
\toprule
\textbf{Map Builder}
& \textbf{GPT-5.5} & \textbf{Sol} & \textbf{Luna} & \textbf{DeepSeek}
& \textbf{Cost} \\
\midrule
\midrule
\textsc{Raw Corpus}
& 65.60 & 69.54 & 63.31 & 68.02
& -- \\
\midrule
GPT-5.5
& \underline{76.60} & \underline{75.31} & \textbf{73.79} & 68.51
& $\leq$\$4,732.14 \\
Sol
& \textbf{77.39} & \textbf{78.03} & \underline{73.60} & \underline{71.06}
& $\leq$\$2,681.76 \\
Luna
& 73.59 & 74.68 & 73.20 & 70.90
& $\leq$\$74.65 \\
DeepSeek
& 73.06 & 73.11 & 69.42 & \textbf{71.11}
& \$310.58 \\
\bottomrule
\end{tabular}
\vspace{-0.1in}
\end{table}

%% file: Figures/fig_evolving_corpora.tex
\begin{figure}[t!]
    \centering
    \includegraphics{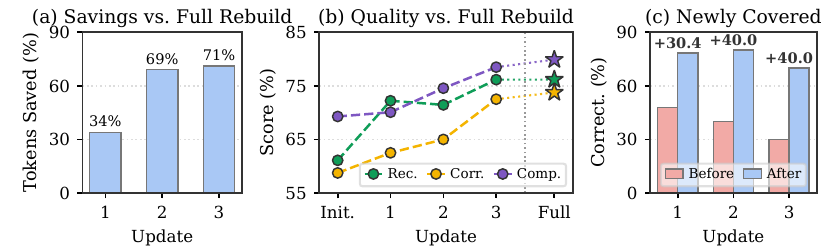}
    \vspace{-0.05in}
    \caption{Incremental map updates on EnterpriseRAG-Bench with GPT-5.5: (a) tokens saved over a full rebuild, (b) quality across map states, and (c) correctness on newly covered questions.}
    \label{fig:evolving_corpora}
    \vspace{-0.1in}
\end{figure}

%% file: Figures/fig_corpus_scale.tex
\begin{figure}[t!]
    \centering
    \includegraphics{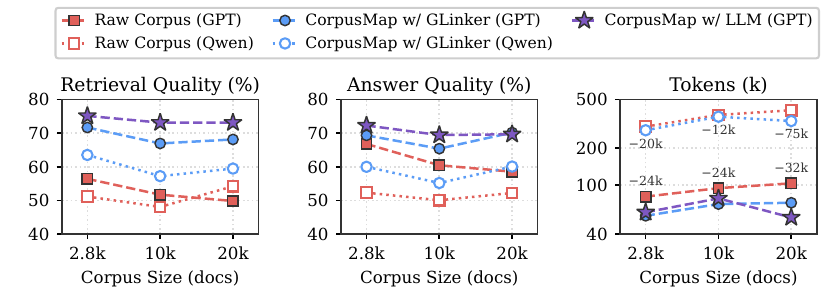}
    \vspace{-0.05in}
    \caption{Corpus scaling on EnterpriseRAG-Bench with GPT-5.6 Luna and Qwen3.8-27B.}
    \label{fig:corpus_scale}
    \vspace{-0.1in}
\end{figure}

%% file: Sections/7_conclusion.tex
\section{Conclusion}
\label{sec:conclusion}

In this work, we introduced \modelname{}, a navigation layer over the corpus anchored on its recurring entities, which is designed to address a practical challenge: access to a large, heterogeneous corpus does not by itself provide guidance on where to search or which sources to inspect.
\modelname{} organizes the corpus around recurring entities by resolving references to the same entity across sources and creating entity-centered representations that link key facts about each entity to the original documents, providing shared anchors that connect related sources across folders and repositories.
Across our experiments, we evaluate 7 models, 3 benchmark datasets, and 5 comparison methods to demonstrate that \modelname{} substantially improves answer quality and evidence discovery while simultaneously reducing per-query token costs.
We envision \modelname{} as a foundation for a broader shift in agentic search over large document collections, from repeatedly searching isolated sources to navigating and reasoning over connected knowledge.

%% file: Sections/8_ai_use_statement.tex
\subsection*{AI use statement}

In this work, we used generative AI tools to assist with implementing and debugging the code for our experiments and with running the experiments and analyzing their outputs.
We have not used generative AI tools for idea proposal or method development, and proof-related tasks are not applicable to this work.
Additionally, we used generative AI tools to assist with editing the paper, including revising text based on the authors' content.
We have reviewed all AI-assisted work and verified the AI-assisted code and analyses against the experimental outputs.
We take responsibility for the final content of this work, including text, claims or artifacts produced with the aid of generative AI.

%% file: Sections/9_ethics_statement.tex
\subsection*{Ethics statement}

Our work aims to enable LLM agents to answer questions whose evidence is distributed across multiple documents of large corpora, such as those of enterprises, and we believe that \modelname{} can contribute to more effective and efficient access to the knowledge scattered across such corpora.
However, we also acknowledge potential risks of our framework.
For example, since \modelname{} consolidates the information about each entity (e.g., a person or a project) from multiple documents into a single Entity Page, it may aggregate private or sensitive information that is otherwise dispersed across the corpus, or expose documents to users who are not permitted to access them.
Also, depending on the underlying corpora and LLMs, the constructed map and the generated answers may contain harmful or biased content.
To address such risks, in real-world deployment, it would be necessary to construct and serve the map in accordance with the access permissions of the corpus (e.g., separately per permission level) and to incorporate safeguards (such as privacy and content filters) for the responsible and safe use of our framework.

%% file: Sections/10_reproducibility.tex
\subsection*{Reproducibility statement}

The details of our experiments are described in \cref{sec:method,sec:experimental_setup}, including the construction protocol of \modelname{} in \cref{alg:entity_map_construction}, and in \cref{sec:appendix_experiment}, which specifies the questions and corpora we use from each benchmark, the evaluation metrics, the agent, and the period of the API calls, while the prompts used to construct \modelname{} are provided in \cref{sec:appendix_prompts}.

%% file: Sections/13_appendix.tex
\section{Additional Experimental Details}
\label[appendix]{sec:appendix_experiment}

\paragraph{Benchmarks and Corpora}
Since \modelname{} targets questions whose evidence is distributed across multiple documents, we select such questions from each benchmark and use the same fixed corpus for every method.
From EnterpriseRAG-Bench~\citep{enterpriseragbench}, we use all 80 questions of the Project Related (40), Conflicting Info (20), and Completeness (20) categories, the only categories in which every question is annotated with at least two gold documents, whereas the other categories include questions annotated with a single gold document (e.g., Basic and Semantic) or with no gold documents (e.g., High Level and Info Not Found).
As the corpus, we use a fixed set of 2,819 documents that includes the gold documents of all questions in the benchmark and the one-hop distractor documents linked to them, and examine larger corpora in \cref{fig:corpus_scale}.
WixQA~\citep{WixQA} provides three splits, ExpertWritten, Simulated, and Synthetic, over a knowledge base of support articles.
We use the 79 questions of ExpertWritten (52) and Simulated (27) that are grounded in more than one article, since the remaining questions, including every Synthetic question, are grounded in a single article, and use the full knowledge base of 6,221 articles as the corpus.
HERB~\citep{HERB} simulates the workspace of a software company, with artifacts such as Slack messages, meeting transcripts, documents, and pull requests, and provides answerable questions of four types.
We use its 238 content-based questions, which ask about information stated across multiple artifacts and are scored against a reference answer.
As the corpus, we use the 6,362 artifacts cited as evidence by these questions, together with three metadata files (e.g., of employees and customers), resulting in 6,365 documents.

\paragraph{Evaluation Metrics}
Following each benchmark, we measure answer quality with LLM judges and retrieval quality with document or context recall, where every LLM-judged metric uses GPT-5.6 Sol, and we examine the robustness of our findings to this choice in \cref{sec:appendix_judge}.
For EnterpriseRAG-Bench, \emph{correctness} is a binary judgment of whether the answer is broadly aligned with the gold answer, addressing the core of the question without conflicting with it, \emph{completeness} is the percentage of the benchmark's atomic answer facts that the judge finds supported by the answer, and \emph{document recall} is the percentage of gold documents included in the supporting set that the agent selects, computed without an LLM.
For WixQA, using its official judge prompts, \emph{factuality} rates how well the answer includes the essential information of the ground-truth answer, and \emph{context recall} rates how well that information is present in the context the agent gathers, namely the outputs of its tool calls together with any candidate list given with the question.
For HERB, following its official evaluator, \emph{content} rates the answer against the reference answer in terms of factual accuracy, completeness, and relevance.
We measure efficiency as the input tokens accumulated over all LLM calls of the agent for a question.
In \cref{tab:main_results}, Overall Quality averages the per-benchmark mean of the quality metrics over the three benchmarks, and Rel. Tok. is the geometric mean over the benchmarks of the input tokens of each method relative to \textsc{Raw Corpus}.

\paragraph{Agent}
Following \citet{enterpriseragbench}, the agent has a fixed execution budget per question, within which it explores the corpus with shell commands (\texttt{ls}, \texttt{tree}, \texttt{find}, \texttt{grep}, \texttt{rg}, \texttt{cat}, \texttt{head}, \texttt{tail}, \texttt{sed}, \texttt{awk}, \texttt{cut}, \texttt{sort}, \texttt{uniq}, \texttt{wc}, \texttt{xargs}, and \texttt{jq}), reads documents, and adds documents to or removes them from its selected supporting set.

\paragraph{LLM Access}
The API calls for the main results in \cref{tab:main_results}, including the construction of each method's artifacts, question answering, and judging, were made between August and September 2026.

\section{Additional Experimental Results}
\label[appendix]{sec:appendix_results}

\input{Tables/tab_enterprise_remaining}
\input{Figures/fig_amortized_cost}

\subsection{Statistical Significance}
\label[appendix]{sec:appendix_significance}

\input{Tables/tab_significance}

To examine whether the improvements of \modelname{} in \cref{tab:main_results} are statistically significant, we perform a paired bootstrap test between \modelname{} and each baseline with each LLM.
Specifically, we first average the score of each question over the three runs of each method, then resample the questions of each benchmark with replacement 100,000 times to recompute the Overall Quality of both methods on the same resampled questions, and correct the resulting $p$-values over the five baselines under each LLM with the Holm--Bonferroni method.
As shown in \cref{tab:significance}, \modelname{} significantly outperforms every baseline with all four LLMs ($p<10^{-4}$), with every 95\% confidence interval lying well above zero.

\subsection{Robustness to the Judge LLM}
\label[appendix]{sec:appendix_judge}

\input{Tables/tab_judge_robustness}

To examine whether our findings depend on the choice of judge, we re-judge the answers of every method with GPT-5.5 in \cref{tab:main_results} using DeepSeek-V4-Pro, an LLM from a different model family than GPT-5.6 Sol, with the same judge prompts, and report the results in \cref{tab:judge_robustness}.
We find that \modelname{} achieves the highest score among the retrieval methods on all five LLM-judged metrics under this judge as well, and that the two judges rank the retrieval methods identically on correctness, completeness, and content.

\subsection{Analysis on Candidate File Paths}
\label[appendix]{sec:appendix_starting_points}

We first describe how the candidates of \modelname{} (\cref{sec:method_navigation}) are constructed.
Given a question, we rank the Entity Pages by BM25 over each page's name, type, overview, key facts, and the names under which the entity appears, and pool the documents linked to the top-ranked pages.
We then rerank the pooled documents by BM25 over their titles and content, and give the top-ranked ones to the agent as candidates, each listed only by its title (if any), file path, and document ID, without any of its content.

\input{Tables/tab_entry_points}

\paragraph{Effect of Candidates}
Since the agent can list and search Entity Pages with the same commands it applies to the raw corpus, it can also find relevant Entity Pages by itself without any candidates; to examine the effect of the candidates, we compare this setting with giving the agent the file paths of candidate Entity Pages, their linked documents, or both, selected by their relevance to the question, and report the results with GPT-5.5 in \cref{tab:entry_points}.
We find that, without candidates, the agent achieves higher overall quality than on \textsc{Raw Corpus} but spends far more tokens exploring the map.
In contrast, each type of candidate substantially improves overall quality, and giving the linked documents (our default) achieves comparable quality with the fewest tokens, even fewer than on \textsc{Raw Corpus}.
Also, replacing the candidates with randomly selected ones of each type lowers overall quality while considerably increasing tokens, indicating that this benefit stems from the relevance of the candidates to the question rather than from simply giving the agent some files in the map.
Meanwhile, on \textsc{Raw Corpus}, giving the agent the file paths of candidate documents selected by their relevance to the question marginally improves overall quality.

\input{Figures/fig_k_sensitivity}

\paragraph{Number of Candidates}
To examine how the number of candidates affects \modelname{}, we vary the number of candidate Entity Pages and of candidate linked documents given to the agent, and report the results with GPT-5.5 in \cref{fig:k_sensitivity}.
We observe that quality improves substantially once more than a single Entity Page or document is given, and then remains relatively stable across a wide range.
Token usage increases mainly when a single document is given, as the agent then explores more by itself to find the documents missing from the candidates, and when a large number of documents are given, as the longer list of candidates is included in the input at every step.

\subsection{Analysis on Entity--Entity Edges}
\label[appendix]{sec:appendix_entity_edges}

\paragraph{Effect of Entity--Entity Edges}
Recall that the map $\gG=(\gE\cup\gD,\gL)$ of \modelname{} is a bipartite graph whose links $\gL$ run between entities and documents (\cref{eq:entity_document_map}), so that any two entities $e$ and $e'$ are connected through every document in $\gN(e)\cap\gN(e')$.
A natural question is whether adding direct edges between entities, as in knowledge graphs, further helps navigation.
To examine this, we add entity--entity edges to the map, either as relations extracted by an LLM or as pairs of entities that share a source document, and list them on each Entity Page.
We first note that these edges do not make new pages reachable: every edge $(e,e')$, of either kind, joins two entities that share a document $d\in\gN(e)\cap\gN(e')$, so it only shortens the existing path $e\to d\to e'$ in $\gG$ to $e\to e'$.
Moreover, search (e.g., \texttt{grep} and \texttt{rg}) already provides a similar shortcut, since the agent can look up the related entities mentioned on each Entity Page; indeed, without entity--entity edges, 86--90\% of the new Entity Pages that the agent opens are found through search.
When these edges are available, the agent follows them in only 11--19\% of questions, and 71\% of these moves lead to no gold document that it has not already found, since each page lists the same neighbors regardless of the question, many of which are irrelevant to it.
Also, even among the gold documents that the agent first reaches through an edge, 79\% are retrieved on the same question without entity--entity edges as well.
As a result, document recall changes by only $-2.6$ to $+0.2$ points, while these edges enlarge Entity Pages by 27--52\% on average and increase input tokens by up to 28\%.
These results suggest that the bipartite map already covers the connections that these edges would add, as they only shorten paths that the agent readily crosses through search, at the cost of larger pages.

\input{Tables/tab_case_study_entity_edges}

\paragraph{Case Study}
We present a case study in \cref{tab:case_study_entity_edges}.
Given the question asking for every internal thread about the rollback loop bug RRB-17, the Entity Page of RRB-17 links four of the five gold documents, and the remaining one, a meeting transcript, is linked from the page of \texttt{installer-rollback-lock}, a configuration flag.
With entity--entity edges, the page of RRB-17 lists an LLM-extracted relation stating that the workaround for RRB-17 deletes \texttt{installer-rollback-lock}, which points the agent to that page.
However, the map already connects the two pages: the source email of the relation is linked from both pages, and each page mentions the other in its text.
Accordingly, without entity--entity edges, the agent reaches the same page through search, since its text mentions RRB-17, and retrieves the same five gold documents.
This example illustrates that the relation provides a shortcut to a page that the map already reaches.

\clearpage
\subsection{Case Study}
\label[appendix]{sec:appendix_case_study}

\input{Tables/tab_case_study}

\clearpage
\section{Prompts}
\label[appendix]{sec:appendix_prompts}
\raggedbottom

In this section, we provide the prompts used in each stage of the \modelname{} construction protocol (\cref{alg:entity_map_construction}).
For Cataloging, the prompts in \cref{fig:prompt_catalog_propose,fig:prompt_catalog_synthesize,fig:prompt_catalog_verify,fig:prompt_catalog_revise} propose a candidate catalog from each set of sampled documents, synthesize these candidates into one, and verify and revise the result, respectively.
For Extraction, the prompt in \cref{fig:prompt_extraction} identifies and types the document-local entities of each document.
For Resolution, the prompt in \cref{fig:prompt_resolution} decides whether each document-local entity is linked to an existing registry entry, added as a new entity, or left unresolved.
For Rendering, the prompt in \cref{fig:prompt_rendering} writes the Entity Page of each retained entity from its linked documents.

\input{Figures/fig_prompt_catalog_propose}
\input{Figures/fig_prompt_catalog_synthesize}
\input{Figures/fig_prompt_catalog_verify}
\input{Figures/fig_prompt_catalog_revise}
\input{Figures/fig_prompt_extraction}
\input{Figures/fig_prompt_resolution}
\input{Figures/fig_prompt_rendering}

%% file: Tables/tab_enterprise_remaining.tex
\begin{table}[h!]
\caption{Results on the remaining questions of EnterpriseRAG-Bench with GPT-5.5. Doc. Rec. is computed over the questions with gold documents.}
\label{tab:enterprise_remaining}
\centering
\small
\renewcommand{\arraystretch}{0.98}
\setlength{\tabcolsep}{5pt}
\begin{tabular}{l c c c c}
\toprule
\textbf{Method}
& \textbf{Doc. Rec.} $\uparrow$
& \textbf{Correct.} $\uparrow$
& \textbf{Complete.} $\uparrow$
& \textbf{Tokens} $\downarrow$ \\
\midrule
\midrule
\textsc{Raw Corpus}
& 89.74 & 89.05 & 86.42 & 250.7k \\
\textbf{\modelname{} (Ours)}
& \textbf{95.38} & \textbf{94.76} & \textbf{89.58} & \textbf{171.7k} \\
\bottomrule
\end{tabular}
\end{table}

%% file: Figures/fig_amortized_cost.tex
\begin{figure}[t!]
    \centering
    \includegraphics[width=0.9\linewidth]{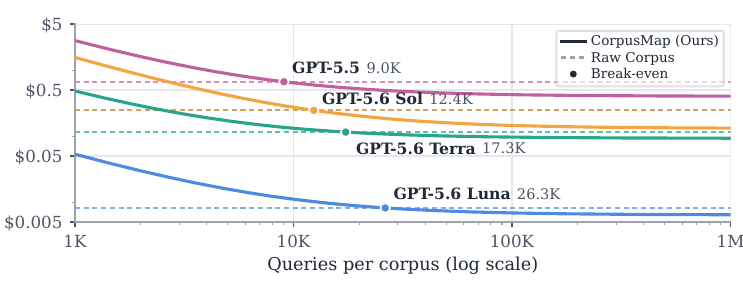}
    \vspace{-0.1in}
    \caption{Cost per query of \modelname{} with amortized construction cost, versus \textsc{Raw Corpus}.}
    \label{fig:amortized_cost}
\end{figure}

%% file: Tables/tab_significance.tex
\begin{table}[h!]
\caption{Gains in Overall Quality of \modelname{} over each baseline in \cref{tab:main_results}, with 95\% confidence intervals from a paired bootstrap over questions. All gains are significant with Holm-corrected $p<10^{-4}$.}
\label{tab:significance}
\vspace{-0.05in}
\centering
\small
\renewcommand{\arraystretch}{0.98}
\setlength{\tabcolsep}{4pt}
\begin{tabular}{l c c c c}
\toprule
\textbf{Baseline} & \textbf{GPT-5.5} & \textbf{Luna} & \textbf{Terra} & \textbf{Sol} \\
\midrule
\midrule
\textsc{Raw Corpus}
& +6.45 {\textsmaller{[4.30, 8.62]}}
& +11.74 {\textsmaller{[9.40, 14.07]}}
& +6.76 {\textsmaller{[4.34, 9.19]}}
& +7.00 {\textsmaller{[5.05, 8.94]}} \\
\textsc{Document Page}
& +6.14 {\textsmaller{[4.05, 8.26]}}
& +9.86 {\textsmaller{[7.63, 12.08]}}
& +9.60 {\textsmaller{[7.28, 11.93]}}
& +8.37 {\textsmaller{[6.24, 10.50]}} \\
\textsc{Group Page}
& +11.48 {\textsmaller{[8.93, 14.01]}}
& +18.03 {\textsmaller{[15.37, 20.70]}}
& +28.67 {\textsmaller{[25.82, 31.48]}}
& +43.46 {\textsmaller{[40.62, 46.26]}} \\
\textsc{LLM Wiki}
& +8.07 {\textsmaller{[5.81, 10.35]}}
& +11.20 {\textsmaller{[8.61, 13.75]}}
& +13.79 {\textsmaller{[11.05, 16.52]}}
& +9.54 {\textsmaller{[7.23, 11.85]}} \\
\textsc{Corpus2Skill}
& +27.07 {\textsmaller{[24.37, 29.80]}}
& +28.32 {\textsmaller{[25.19, 31.41]}}
& +28.65 {\textsmaller{[25.69, 31.61]}}
& +30.25 {\textsmaller{[27.38, 33.12]}} \\
\bottomrule
\end{tabular}
\end{table}

%% file: Tables/tab_judge_robustness.tex
\begin{table}[h!]
\caption{LLM-judged metrics of the GPT-5.5 answers in \cref{tab:main_results}, judged by DeepSeek-V4-Pro instead of GPT-5.6 Sol. The last row reports Kendall's $\tau$ between the rankings of the retrieval methods under the two judges.}
\label{tab:judge_robustness}
\vspace{-0.05in}
\centering
\small
\renewcommand{\arraystretch}{0.98}
\ADLactivate
\setlength{\tabcolsep}{4pt}
\begin{tabular}{l c c c c c}
\toprule
& \multicolumn{2}{c}{\textbf{EnterpriseRAG-Bench}}
& \multicolumn{2}{c}{\textbf{WixQA}}
& \textbf{HERB} \\
\cmidrule(lr){2-3} \cmidrule(lr){4-5} \cmidrule(lr){6-6}
\textbf{Method}
& \textbf{Correct.} & \textbf{Complete.}
& \textbf{Ctx. Rec.} & \textbf{Fact.}
& \textbf{Content} \\
\midrule
\midrule
\textsc{Raw Corpus} & \underline{83.75} & \underline{78.11} & 50.95 & 67.09 & 65.04 \\
\textsc{Document Page} & 83.33 & 73.61 & 58.12 & \underline{67.72} & \underline{65.56} \\
\textsc{Group Page} & 76.25 & 68.90 & 35.13 & 62.55 & 65.53 \\
\textsc{LLM Wiki} & 80.42 & 73.10 & 58.54 & 67.09 & 64.45 \\
\textsc{Corpus2Skill} & 67.50 & 61.76 & \underline{60.55} & 61.71 & 29.72 \\
\textbf{\modelname{} (Ours)} & \textbf{95.00} & \textbf{83.24} & \textbf{67.62} & \textbf{69.83} & \textbf{67.29} \\
\noalign{\vskip 0.25ex}\cdashline{1-6}\noalign{\vskip 0.75ex}
\textit{Gold Documents (Oracle)} & \textit{97.08} & \textit{80.86} & \textit{73.73} & \textit{76.79} & \textit{70.85} \\
\midrule
Kendall's $\tau$ & 1.00 & 1.00 & 0.47 & 0.83 & 1.00 \\
\bottomrule
\end{tabular}
\end{table}

%% file: Tables/tab_entry_points.tex
\begin{table}[t!]
\caption{Overall Quality with candidate file paths on EnterpriseRAG-Bench with GPT-5.5.}
\label{tab:entry_points}
\centering
\small
\renewcommand{\arraystretch}{0.98}
\ADLactivate
\setlength{\tabcolsep}{3pt}
\begin{tabular}{c l c c c c}
\toprule
& & \multicolumn{2}{c}{\textbf{Relevant}}
& \multicolumn{2}{c}{\textbf{Random}} \\
\cmidrule(lr){3-4} \cmidrule(lr){5-6}
& \textbf{Method}
& \textbf{Quality} & \textbf{Tokens}
& \textbf{Quality} & \textbf{Tokens} \\
\midrule
\midrule
& \textsc{Raw Corpus}
& 65.60 & 206.5k
& -- & -- \\
& \textsc{Raw Corpus} w/ Candidates
& 66.19 & 107.4k
& -- & -- \\
\midrule
\multirow{4}{*}{\rotatebox{90}{\scriptsize\textbf{\modelname{}}}}
& w/o Candidates
& 69.04 & 496.4k
& -- & -- \\
\noalign{\vskip 0.25ex}\cdashline{2-6}\noalign{\vskip 0.75ex}
& w/ Candidate Entity Pages
& \underline{76.68} & 247.1k
& 68.65 & 404.6k \\
& w/ Candidate Entity-Linked Docs
& 76.60 & 88.1k
& 60.76 & 413.5k \\
& w/ Both Candidates
& \textbf{77.43} & 102.7k
& 62.30 & 349.1k \\
\bottomrule
\end{tabular}
\ADLinactivate
\end{table}

%% file: Figures/fig_k_sensitivity.tex
\begin{figure}[t!]
    \centering
    \includegraphics{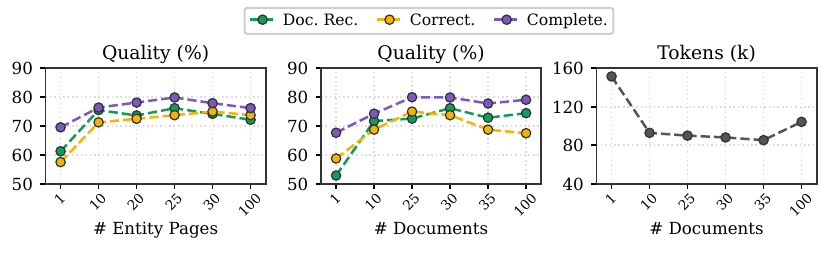}
    \caption{Results with varying the number of candidate Entity Pages (left) and of candidate linked documents (middle and right) while fixing the other, on EnterpriseRAG-Bench with GPT-5.5.}
    \label{fig:k_sensitivity}
\end{figure}

%% file: Tables/tab_case_study_entity_edges.tex
\begin{table*}[h!]
\centering
\caption{Case study of entity--entity edges on EnterpriseRAG-Bench. Blue and orange boxes denote documents and Entity Pages, as in \cref{fig:concept}.}
\label{tab:case_study_entity_edges}
\small
\renewcommand{\arraystretch}{1.3}
\setlength{\tabcolsep}{5pt}
\begin{tabular}{>{\raggedright\arraybackslash}p{0.155\linewidth} >{\raggedright\arraybackslash}p{0.785\linewidth}}
\toprule
\rowcolor{gray!8}
\textbf{Question} & List every internal communication thread (email, Slack, and meeting notes) about the Redwood Private upgrade `rollback loop' bug (including references to RRB-17 or `stuck rollback'). \\
\midrule
\textbf{Gold Documents} & \docchip{D1: \#eng thread (Slack)} \ \docchip{D2: INC-2147 thread (Slack)} \ \docchip{D3: customer email (Gmail)} \newline
\docchip{D4: RRB-17 root cause and patch plan (Gmail)} \ \docchip{D5: escalation meeting (Fireflies)} \\
\midrule
\textbf{Entity--Document} & \entchip{RRB-17} links D1--D4, and \entchip{installer-rollback-lock} links D1--D5. \newline
\textbf{Path:} \entchip{RRB-17} $\rightarrow$ \docchip{D4} $\rightarrow$ \entchip{installer-rollback-lock} \\
\midrule
\textbf{Entity--Entity} & \textbf{Path:} \entchip{RRB-17} $\rightarrow$ \texttt{workaround deletes} $\rightarrow$ \entchip{installer-rollback-lock} \newline
Extracted from D4: ``Workaround (current): delete CM \texttt{installer-rollback-lock} and restart installer-controller.'' \\
\bottomrule
\end{tabular}
\end{table*}

%% file: Tables/tab_case_study.tex
\begin{table*}[h!]
\centering
\caption{Case study comparing \modelname{} with \textsc{Corpus2Skill} on EnterpriseRAG-Bench. Blue and orange boxes denote documents and Entity Pages, as in \cref{fig:concept}.}
\label{tab:case_study_streaming}
\small
\renewcommand{\arraystretch}{1.3}
\setlength{\tabcolsep}{5pt}
\begin{tabular}{>{\raggedright\arraybackslash}p{0.155\linewidth} >{\raggedright\arraybackslash}p{0.785\linewidth}}
\toprule
\rowcolor{gray!8}
\textbf{Question} & In the Deterministic Playback Manifest v1, how is manifest signing/integrity represented (signature vs.\ integrity fields)? \\
\midrule
\textbf{Gold Answer} & In v1, the manifest does not embed a \texttt{signature} blob; integrity is expressed via an optional \texttt{integrity} field and an \texttt{integrity\_ref} URI. The earlier draft instead listed an embedded \texttt{signature} field. \newline
\textbf{Gold documents:} \docchip{D1: manifest draft (Google Drive)} \ \docchip{D2: manifest v1 (Confluence)} \\
\midrule
\cellcolor{gray!8}\textbf{\textsc{Corpus2Skill}} &
\textbf{Explored:} ROOT $\rightarrow$ all four top-level skills $\rightarrow$ several clusters below them; D1 and D2 both sit under one cluster that the agent does not open. \newline
\textbf{Lookups:} ``Deterministic Playback'' (no match), ``manifest'' (an unrelated cluster), ``integrity'' (no match). \newline
{\color{failtext}\textbf{Answer:} ``I couldn't locate a document for `Deterministic Playback Manifest v1' in the available corpus.''} \ \redx \\
\midrule
\cellcolor{walkfill}\textbf{\modelname{} (Ours)} &
\textbf{Path:} \entchip{ENG-8192} $\rightarrow$ \docchip{D1} $\rightarrow$ \entchip{Serving Runtime} $\rightarrow$ \docchip{D2} \newline
(1) Searches Entity Pages for ``playback'' and opens \entchip{ENG-8192}, whose linked documents include D1. \newline
(2) Reads \docchip{D1}: ``signature: optional signed blob for integrity verification.'' \newline
(3) Searches Entity Pages with these terms and opens \entchip{Serving Runtime}, whose linked documents include D2. \newline
(4) Reads \docchip{D2}: ``\texttt{signature} is no longer a direct embedded blob in v1.'' \newline
{\color{walktext}\textbf{Answer:} v1 uses an optional \texttt{integrity} field and an \texttt{integrity\_ref} URI rather than an embedded signature.} \ \greencheck \\
\bottomrule
\end{tabular}
\end{table*}

%% file: Figures/fig_prompt_catalog_propose.tex
\begin{figure}[ht!]
\centering
\scriptsize
\begin{tcolorbox}[
  title=Prompt for proposing a candidate entity type catalog in the Cataloging stage,
  fonttitle=\bfseries,
  rounded corners,
  width=\textwidth
]
You are given raw documents sampled from multiple source types in an enterprise corpus. \\

Discover candidate entity types for distinct, identifiable subjects supported by these documents that could recur across documents and serve as navigation anchors. \\

An entity type categorizes independently referenceable subjects, not standalone facts, proposed actions, or unnamed generic descriptions. \\
Named artifacts, events, or rules qualify only when they have a stable identifier or durable name; exclude ad hoc conditions and key-value assignments. \\
Treat entity types as identity schemas, not topical categories. \\
A type is valid only when all its entities can be recognized, distinguished, and canonicalized using the same kind of evidence. \\
Related subjects that require different identity rules must use separate types. \\
Do not create entity types solely for input document records or their source identifiers; document records belong to the document layer unless they are independently referenced as subjects across documents. \\
The set of types is open. Do not assume a predefined ontology or a target number of types. \\
Ground every type in exact entity strings observed in the documents. \\
Treat observed\_entities as validated evidence, not illustrative examples. \\
Every observed entity must itself be a stable name or identifier for the subject, not a statement about the subject or its current state or value. \\
Before returning, first check for omitted valid identity schemas. Then verify that every observed entity is the same kind of subject, satisfies the type definition, and can be canonicalized using the same identity criteria. Otherwise split the type or remove the outlier. \\

Return JSON only in the following format: \\

{\ttfamily\frenchspacing
\{ \\
\hspace*{1em}"entity\_types": [ \\
\hspace*{2em}\{ \\
\hspace*{3em}"name": "...", \\
\hspace*{3em}"definition": "...", \\
\hspace*{3em}\parbox[t]{\dimexpr\linewidth-3em\relax}{"identity\_criteria": "How instances are identified, distinguished, and canonicalized.",} \\
\hspace*{3em}"observed\_entities": ["...", "..."] \\
\hspace*{2em}\} \\
\hspace*{1em}] \\
\} \\
}

Raw documents: \\
\{raw\_documents\}
\end{tcolorbox}
\caption{Prompt for proposing a candidate entity type catalog in the Cataloging stage. \{\} indicates a placeholder, filled with the documents of one sampled set grouped by their source.}
\label{fig:prompt_catalog_propose}
\end{figure}

%% file: Figures/fig_prompt_catalog_synthesize.tex
{\scriptsize
\begin{tcolorbox}[
  breakable,
  title=Prompt for synthesizing the candidate catalogs in the Cataloging stage,
  fonttitle=\bfseries,
  rounded corners,
  width=\textwidth
]
You are synthesizing one stable, query-independent Entity Type Catalog from independently discovered source catalogs. \\

Treat every source Type as a proposal, not as ground truth. Produce identity schemas for distinct, independently referenceable subjects that can recur across enterprise documents and serve as durable navigation anchors. \\

Required decision order for every source Type: \\
\makebox[1.2em][l]{1.}\parbox[t]{\dimexpr\linewidth-1.2em\relax}{Remove only individually invalid observations using the narrow observation exclusion rules.\strut} \\
\makebox[1.2em][l]{2.}\parbox[t]{\dimexpr\linewidth-1.2em\relax}{Narrow an overbroad proposal to a coherent identity schema when possible.\strut} \\
\makebox[1.2em][l]{3.}\parbox[t]{\dimexpr\linewidth-1.2em\relax}{Split a mixed proposal across materially different identity schemas.\strut} \\
\makebox[1.2em][l]{4.}\parbox[t]{\dimexpr\linewidth-1.2em\relax}{Merge or remap an equivalent, narrower, broader, or subsumed proposal to existing final Types.\strut} \\
\makebox[1.2em][l]{5.}\parbox[t]{\dimexpr\linewidth-1.2em\relax}{Reject only after all earlier steps fail, every source observation is invalid, no coherent stable identity schema can be defined, and no safe mapping or split to existing final Types exists.\strut} \\[\baselineskip]
A valid source identity schema must never be rejected to hide overlap, granularity, criteria, or identity defects. Do not optimize for a target number of final Types. \\

Important constraints: \\
\makebox[1.2em][l]{-}\parbox[t]{\dimexpr\linewidth-1.2em\relax}{Account for every supplied \mbox{source\_type\_id} exactly once in \mbox{source\_type\_mapping}.\strut} \\
\makebox[1.2em][l]{-}\parbox[t]{\dimexpr\linewidth-1.2em\relax}{A mapped source Type maps to exactly one final Type ID.\strut} \\
\makebox[1.2em][l]{-}\parbox[t]{\dimexpr\linewidth-1.2em\relax}{A split source Type maps to two or more final Type IDs.\strut} \\
\makebox[1.2em][l]{-}\parbox[t]{\dimexpr\linewidth-1.2em\relax}{A rejected source Type maps to no final Type IDs.\strut} \\
\makebox[1.2em][l]{-}\parbox[t]{\dimexpr\linewidth-1.2em\relax}{\mbox{rejected\_source\_types} is required and must be [] when no source Type is rejected. It is an audit tombstone; never erase a source Type or its mapping.\strut} \\
\makebox[1.2em][l]{-}\parbox[t]{\dimexpr\linewidth-1.2em\relax}{Every rejected mapping has exactly one tombstone, and every tombstone binds exactly one rejected mapping. Tombstones are forbidden for mapped or split source Types.\strut} \\
\makebox[1.2em][l]{-}\parbox[t]{\dimexpr\linewidth-1.2em\relax}{A tombstone must use \mbox{reason\_code} \mbox{all\_observations\_invalid\_no\_stable\_identity\_or\_safe\_remap}, assert \mbox{valid\_observation\_count} 0 and \mbox{stable\_identity\_schema} false, provide nonempty reason, \mbox{identity\_assessment}, and \mbox{remap\_assessment}, and use an empty \mbox{remap\_candidate\_final\_type\_ids} array.\strut} \\
\makebox[1.2em][l]{-}\parbox[t]{\dimexpr\linewidth-1.2em\relax}{\mbox{observation\_audit} must reproduce every exact \mbox{observed\_entities} value from the rejected source Type exactly once and in source order. Each item must use the exact \mbox{observed\_entity}, a concrete reason, and one of the narrow reason codes \mbox{missing\_stable\_identifier\_or\_provenance} or \mbox{non\_atomic\_compound\_observation}.\strut} \\
\makebox[1.2em][l]{-}\parbox[t]{\dimexpr\linewidth-1.2em\relax}{\mbox{excluded\_source\_observations} is forbidden for rejected source Types.\strut} \\
\makebox[1.2em][l]{-}\parbox[t]{\dimexpr\linewidth-1.2em\relax}{Every final Type must be supported by at least one mapped or split source Type.\strut} \\
\makebox[1.2em][l]{-}\parbox[t]{\dimexpr\linewidth-1.2em\relax}{Do not invent \mbox{observed\_entities}. Every string must occur exactly in a supplied source catalog and be supported by a mapped or split source Type.\strut} \\
\makebox[1.2em][l]{-}\parbox[t]{\dimexpr\linewidth-1.2em\relax}{An individually noisy observation may be excluded only from a mapped source Type when it lacks stable identity/provenance or is a non-atomic compound.\strut} \\
\makebox[1.2em][l]{-}\parbox[t]{\dimexpr\linewidth-1.2em\relax}{Record each individual exclusion with exact \mbox{source\_type\_id}, exact \mbox{observed\_entity}, and a concrete reason. Never use an exclusion to hide a mapping, overlap, granularity, criteria, identity, or invention defect.\strut} \\
\makebox[1.2em][l]{-}\parbox[t]{\dimexpr\linewidth-1.2em\relax}{\mbox{inclusion\_criteria} and \mbox{exclusion\_criteria} must be semantic, consistent criteria. \mbox{identity\_criteria} must explain reliable identification, distinction, and canonicalization.\strut} \\
\makebox[1.2em][l]{-}\parbox[t]{\dimexpr\linewidth-1.2em\relax}{Use stable lowercase snake-case IDs matching \mbox{type\_\textless{}name\textgreater{}}. Type names must be unique.\strut} \\
\makebox[1.2em][l]{-}\parbox[t]{\dimexpr\linewidth-1.2em\relax}{Return only JSON.\strut} \\[\baselineskip]
Return exactly this structure: \\

{\ttfamily\frenchspacing
\{ \\
\hspace*{1em}\parbox[t]{\dimexpr\linewidth-1em\relax}{"source\_type\_mapping": [} \\
\hspace*{2em}\parbox[t]{\dimexpr\linewidth-2em\relax}{\{} \\
\hspace*{3em}\parbox[t]{\dimexpr\linewidth-3em\relax}{"source\_type\_id": "seed42\_v3::type\_0001",} \\
\hspace*{3em}\parbox[t]{\dimexpr\linewidth-3em\relax}{"resolution": "mapped",} \\
\hspace*{3em}\parbox[t]{\dimexpr\linewidth-3em\relax}{"final\_type\_ids": ["type\_example"],} \\
\hspace*{3em}\parbox[t]{\dimexpr\linewidth-3em\relax}{"reason": "Short evidence-based mapping reason."} \\
\hspace*{2em}\parbox[t]{\dimexpr\linewidth-2em\relax}{\},} \\
\hspace*{2em}\parbox[t]{\dimexpr\linewidth-2em\relax}{\{} \\
\hspace*{3em}\parbox[t]{\dimexpr\linewidth-3em\relax}{"source\_type\_id": "seed43\_v3::type\_0001",} \\
\hspace*{3em}\parbox[t]{\dimexpr\linewidth-3em\relax}{"resolution": "rejected",} \\
\hspace*{3em}\parbox[t]{\dimexpr\linewidth-3em\relax}{"final\_type\_ids": [],} \\
\hspace*{3em}\parbox[t]{\dimexpr\linewidth-3em\relax}{"reason": "Why the strict rejection rule is satisfied."} \\
\hspace*{2em}\parbox[t]{\dimexpr\linewidth-2em\relax}{\}} \\
\hspace*{1em}\parbox[t]{\dimexpr\linewidth-1em\relax}{],} \\
\hspace*{1em}\parbox[t]{\dimexpr\linewidth-1em\relax}{"rejected\_source\_types": [} \\
\hspace*{2em}\parbox[t]{\dimexpr\linewidth-2em\relax}{\{} \\
\hspace*{3em}\parbox[t]{\dimexpr\linewidth-3em\relax}{"source\_type\_id": "seed43\_v3::type\_0001",} \\
\hspace*{3em}\parbox[t]{\dimexpr\linewidth-3em\relax}{"reason\_code": "all\_observations\_invalid\_no\_stable\_identity\_or\_safe\_remap",} \\
\hspace*{3em}\parbox[t]{\dimexpr\linewidth-3em\relax}{"reason": "Concrete type-level rejection reason.",} \\
\hspace*{3em}\parbox[t]{\dimexpr\linewidth-3em\relax}{"valid\_observation\_count": 0,} \\
\hspace*{3em}\parbox[t]{\dimexpr\linewidth-3em\relax}{"stable\_identity\_schema": false,} \\
\hspace*{3em}\parbox[t]{\dimexpr\linewidth-3em\relax}{"identity\_assessment": "Why narrowing or splitting cannot produce a coherent stable identity schema.",} \\
\hspace*{3em}\parbox[t]{\dimexpr\linewidth-3em\relax}{"remap\_assessment": "Why merging, mapping, or splitting to final Types is unsafe.",} \\
\hspace*{3em}\parbox[t]{\dimexpr\linewidth-3em\relax}{"remap\_candidate\_final\_type\_ids": [],} \\
\hspace*{3em}\parbox[t]{\dimexpr\linewidth-3em\relax}{"observation\_audit": [} \\
\hspace*{4em}\parbox[t]{\dimexpr\linewidth-4em\relax}{\{} \\
\hspace*{5em}\parbox[t]{\dimexpr\linewidth-5em\relax}{"observed\_entity": "Exact source observation.",} \\
\hspace*{5em}\parbox[t]{\dimexpr\linewidth-5em\relax}{"reason\_code": "missing\_stable\_identifier\_or\_provenance",} \\
\hspace*{5em}\parbox[t]{\dimexpr\linewidth-5em\relax}{"reason": "Concrete reason this exact observation is invalid."} \\
\hspace*{4em}\parbox[t]{\dimexpr\linewidth-4em\relax}{\}} \\
\hspace*{3em}\parbox[t]{\dimexpr\linewidth-3em\relax}{]} \\
\hspace*{2em}\parbox[t]{\dimexpr\linewidth-2em\relax}{\}} \\
\hspace*{1em}\parbox[t]{\dimexpr\linewidth-1em\relax}{],} \\
\hspace*{1em}\parbox[t]{\dimexpr\linewidth-1em\relax}{"excluded\_source\_observations": [} \\
\hspace*{2em}\parbox[t]{\dimexpr\linewidth-2em\relax}{\{} \\
\hspace*{3em}\parbox[t]{\dimexpr\linewidth-3em\relax}{"source\_type\_id": "seed42\_v3::type\_0001",} \\
\hspace*{3em}\parbox[t]{\dimexpr\linewidth-3em\relax}{"observed\_entity": "Exact noisy source example.",} \\
\hspace*{3em}\parbox[t]{\dimexpr\linewidth-3em\relax}{"reason": "Why this exact observation is individually invalid."} \\
\hspace*{2em}\parbox[t]{\dimexpr\linewidth-2em\relax}{\}} \\
\hspace*{1em}\parbox[t]{\dimexpr\linewidth-1em\relax}{],} \\
\hspace*{1em}\parbox[t]{\dimexpr\linewidth-1em\relax}{"final\_types": [} \\
\hspace*{2em}\parbox[t]{\dimexpr\linewidth-2em\relax}{\{} \\
\hspace*{3em}\parbox[t]{\dimexpr\linewidth-3em\relax}{"type\_id": "type\_example",} \\
\hspace*{3em}\parbox[t]{\dimexpr\linewidth-3em\relax}{"name": "Example Type",} \\
\hspace*{3em}\parbox[t]{\dimexpr\linewidth-3em\relax}{"definition": "A precise definition of the subject kind.",} \\
\hspace*{3em}\parbox[t]{\dimexpr\linewidth-3em\relax}{"inclusion\_criteria": [} \\
\hspace*{4em}\parbox[t]{\dimexpr\linewidth-4em\relax}{"Semantic evidence that qualifies a subject for this Type."} \\
\hspace*{3em}\parbox[t]{\dimexpr\linewidth-3em\relax}{],} \\
\hspace*{3em}\parbox[t]{\dimexpr\linewidth-3em\relax}{"exclusion\_criteria": [} \\
\hspace*{4em}\parbox[t]{\dimexpr\linewidth-4em\relax}{"A confusable subject that does not qualify for this Type."} \\
\hspace*{3em}\parbox[t]{\dimexpr\linewidth-3em\relax}{],} \\
\hspace*{3em}\parbox[t]{\dimexpr\linewidth-3em\relax}{"identity\_criteria": "How instances are reliably identified, distinguished, and canonicalized.",} \\
\hspace*{3em}\parbox[t]{\dimexpr\linewidth-3em\relax}{"observed\_entities": [} \\
\hspace*{4em}\parbox[t]{\dimexpr\linewidth-4em\relax}{"Exact positive example copied from a supplied source catalog."} \\
\hspace*{3em}\parbox[t]{\dimexpr\linewidth-3em\relax}{]} \\
\hspace*{2em}\parbox[t]{\dimexpr\linewidth-2em\relax}{\}} \\
\hspace*{1em}\parbox[t]{\dimexpr\linewidth-1em\relax}{]} \\
\} \\
}

Source catalog snapshot: \\
\{synthesis\_input\}
\end{tcolorbox}
}
\captionof{figure}{Prompt for synthesizing the candidate catalogs in the Cataloging stage. \{\} indicates a placeholder, filled with the candidate catalogs proposed from all sampled sets.}
\label{fig:prompt_catalog_synthesize}

%% file: Figures/fig_prompt_catalog_verify.tex
{\scriptsize
\begin{tcolorbox}[
  breakable,
  title=Prompt for verifying the synthesized catalog in the Cataloging stage,
  fonttitle=\bfseries,
  rounded corners,
  width=\textwidth
]
You are the independent adversarial verifier for a synthesized Entity Type Catalog. You did not create the draft. Do not rewrite it or propose a replacement. Find concrete defects and counterexamples. \\

Inputs are immutable source Catalog snapshots, one synthesized draft, and SHA-256 digests binding both. \\

Verification duties, in order: \\
\makebox[1.2em][l]{1.}\parbox[t]{\dimexpr\linewidth-1.2em\relax}{Audit individual observation exclusions.\strut} \\
\makebox[1.2em][l]{2.}\parbox[t]{\dimexpr\linewidth-1.2em\relax}{Test whether each source proposal can be narrowed to a coherent identity.\strut} \\
\makebox[1.2em][l]{3.}\parbox[t]{\dimexpr\linewidth-1.2em\relax}{Test whether it can be split, merged, mapped, or remapped safely.\strut} \\
\makebox[1.2em][l]{4.}\parbox[t]{\dimexpr\linewidth-1.2em\relax}{Only then audit any rejected source Type.\strut} \\
\makebox[1.2em][l]{5.}\parbox[t]{\dimexpr\linewidth-1.2em\relax}{Check coverage, overlap, granularity, criteria consistency, identity criteria, observed examples, and unsupported invention across final Types.\strut} \\[\baselineskip]
For every rejected source Type, adversarially verify all of the following: \\
\makebox[1.2em][l]{-}\parbox[t]{\dimexpr\linewidth-1.2em\relax}{exactly one rejected mapping and exactly one tombstone exist;\strut} \\
\makebox[1.2em][l]{-}\parbox[t]{\dimexpr\linewidth-1.2em\relax}{the tombstone covers every exact source \mbox{observed\_entities} value exactly once and in source order;\strut} \\
\makebox[1.2em][l]{-}\parbox[t]{\dimexpr\linewidth-1.2em\relax}{every observation is genuinely invalid under its narrow reason code;\strut} \\
\makebox[1.2em][l]{-}\parbox[t]{\dimexpr\linewidth-1.2em\relax}{valid\_observation\_count is 0 and \mbox{stable\_identity\_schema} is false;\strut} \\
\makebox[1.2em][l]{-}\parbox[t]{\dimexpr\linewidth-1.2em\relax}{no coherent stable source identity schema can be defined by narrowing or splitting;\strut} \\
\makebox[1.2em][l]{-}\parbox[t]{\dimexpr\linewidth-1.2em\relax}{no safe map, remap, merge, or split to existing final Types exists;\strut} \\
\makebox[1.2em][l]{-}\parbox[t]{\dimexpr\linewidth-1.2em\relax}{identity\_assessment and \mbox{remap\_assessment} are concrete and complete;\strut} \\
\makebox[1.2em][l]{-}\parbox[t]{\dimexpr\linewidth-1.2em\relax}{remap\_candidate\_final\_type\_ids is empty;\strut} \\
\makebox[1.2em][l]{-}\parbox[t]{\dimexpr\linewidth-1.2em\relax}{no \mbox{excluded\_source\_observations} item belongs to the rejected source Type;\strut} \\
\makebox[1.2em][l]{-}\parbox[t]{\dimexpr\linewidth-1.2em\relax}{rejection is not hiding a coverage, overlap, granularity, criteria, or identity defect.\strut} \\[\baselineskip]
Emit a \mbox{coverage\_violations} item for that exact \mbox{source\_type\_id} if any source observation is a valid identity, a coherent source identity schema exists, a safe mapping or split exists, the audit is incomplete or inconsistent, or rejection hides another defect. A rejection coverage finding is always blocking and must never receive a policy annotation. \\

Other duties: \\
\makebox[1.2em][l]{-}\parbox[t]{\dimexpr\linewidth-1.2em\relax}{Flag final Type pairs that accept the same subject without a reliable evidence-based distinction.\strut} \\
\makebox[1.2em][l]{-}\parbox[t]{\dimexpr\linewidth-1.2em\relax}{Flag incompatible flat parent/child granularity and internally mixed identity schemas.\strut} \\
\makebox[1.2em][l]{-}\parbox[t]{\dimexpr\linewidth-1.2em\relax}{Flag contradictory or brittle inclusion/exclusion criteria.\strut} \\
\makebox[1.2em][l]{-}\parbox[t]{\dimexpr\linewidth-1.2em\relax}{Flag Types that cannot be reliably identified, distinguished, and canonicalized with one coherent kind of evidence.\strut} \\
\makebox[1.2em][l]{-}\parbox[t]{\dimexpr\linewidth-1.2em\relax}{Flag observed entities that do not fit and invented final Types.\strut} \\[\baselineskip]
Policy annotations retain the following narrow exceptions: \\
\makebox[1.2em][l]{-}\parbox[t]{\dimexpr\linewidth-1.2em\relax}{Omit an annotation when a finding is hard or eligibility is uncertain.\strut} \\
\makebox[1.2em][l]{-}\parbox[t]{\dimexpr\linewidth-1.2em\relax}{Multi-Type overlap or granularity may be \mbox{soft\_warning} only with exact, distinct, scalar runtime-visible route values in definition or \mbox{identity\_criteria}. Single-Type granularity is never soft.\strut} \\
\makebox[1.2em][l]{-}\parbox[t]{\dimexpr\linewidth-1.2em\relax}{observed\_entity\_failure may be \mbox{audited\_exclusion\_candidate} only for one exact observation from one mapped, non-split source Type that lacks stable identity/provenance or is a non-atomic compound.\strut} \\
\makebox[1.2em][l]{-}\parbox[t]{\dimexpr\linewidth-1.2em\relax}{Audit each existing \mbox{excluded\_source\_observations} item with \mbox{excluded\_source\_observations}/\textless{}index\textgreater{} and \mbox{audited\_exclusion} under the same narrow rule.\strut} \\
\makebox[1.2em][l]{-}\parbox[t]{\dimexpr\linewidth-1.2em\relax}{\raggedright The only observation exclusion reason codes are \mbox{missing\_stable\_identifier\_or\_provenance} and \mbox{non\_atomic\_compound\_observation}.\strut} \\
\makebox[1.2em][l]{-}\parbox[t]{\dimexpr\linewidth-1.2em\relax}{Do not annotate rejected Type tombstones or exclusions from rejected/split source Types.\strut} \\[\baselineskip]
Rules: \\
\makebox[1.2em][l]{-}\parbox[t]{\dimexpr\linewidth-1.2em\relax}{Do not approve on plausibility; actively seek breaking examples.\strut} \\
\makebox[1.2em][l]{-}\parbox[t]{\dimexpr\linewidth-1.2em\relax}{Do not report style preferences or rewrite Types.\strut} \\
\makebox[1.2em][l]{-}\parbox[t]{\dimexpr\linewidth-1.2em\relax}{Every violation references supplied source or final Type IDs.\strut} \\
\makebox[1.2em][l]{-}\parbox[t]{\dimexpr\linewidth-1.2em\relax}{verdict is fail when any violation array is nonempty and pass only when all are empty.\strut} \\
\makebox[1.2em][l]{-}\parbox[t]{\dimexpr\linewidth-1.2em\relax}{Echo both supplied hashes exactly.\strut} \\
\makebox[1.2em][l]{-}\parbox[t]{\dimexpr\linewidth-1.2em\relax}{Return only JSON.\strut} \\

Return exactly this structure: \\

{\ttfamily\frenchspacing
\{ \\
\hspace*{1em}\parbox[t]{\dimexpr\linewidth-1em\relax}{"synthesis\_input\_sha256": "\{synthesis\_input\_sha256\}",} \\
\hspace*{1em}\parbox[t]{\dimexpr\linewidth-1em\relax}{"synthesis\_sha256": "\{synthesis\_sha256\}",} \\
\hspace*{1em}\parbox[t]{\dimexpr\linewidth-1em\relax}{"coverage\_violations": [} \\
\hspace*{2em}\parbox[t]{\dimexpr\linewidth-2em\relax}{\{} \\
\hspace*{3em}\parbox[t]{\dimexpr\linewidth-3em\relax}{"source\_type\_id": "seed42\_v3::type\_0001",} \\
\hspace*{3em}\parbox[t]{\dimexpr\linewidth-3em\relax}{"problem": "Concrete coverage or rejection-audit defect."} \\
\hspace*{2em}\parbox[t]{\dimexpr\linewidth-2em\relax}{\}} \\
\hspace*{1em}\parbox[t]{\dimexpr\linewidth-1em\relax}{],} \\
\hspace*{1em}\parbox[t]{\dimexpr\linewidth-1em\relax}{"overlap\_violations": [} \\
\hspace*{2em}\parbox[t]{\dimexpr\linewidth-2em\relax}{\{} \\
\hspace*{3em}\parbox[t]{\dimexpr\linewidth-3em\relax}{"left\_type\_id": "type\_left",} \\
\hspace*{3em}\parbox[t]{\dimexpr\linewidth-3em\relax}{"right\_type\_id": "type\_right",} \\
\hspace*{3em}\parbox[t]{\dimexpr\linewidth-3em\relax}{"counterexample": "Exact confusable observed entity string.",} \\
\hspace*{3em}\parbox[t]{\dimexpr\linewidth-3em\relax}{"problem": "Why both Types accept the same subject."} \\
\hspace*{2em}\parbox[t]{\dimexpr\linewidth-2em\relax}{\}} \\
\hspace*{1em}\parbox[t]{\dimexpr\linewidth-1em\relax}{],} \\
\hspace*{1em}\parbox[t]{\dimexpr\linewidth-1em\relax}{"granularity\_violations": [} \\
\hspace*{2em}\parbox[t]{\dimexpr\linewidth-2em\relax}{\{} \\
\hspace*{3em}\parbox[t]{\dimexpr\linewidth-3em\relax}{"type\_ids": ["type\_parent", "type\_child"],} \\
\hspace*{3em}\parbox[t]{\dimexpr\linewidth-3em\relax}{"problem": "Why these Types have incompatible flat granularity."} \\
\hspace*{2em}\parbox[t]{\dimexpr\linewidth-2em\relax}{\}} \\
\hspace*{1em}\parbox[t]{\dimexpr\linewidth-1em\relax}{],} \\
\hspace*{1em}\parbox[t]{\dimexpr\linewidth-1em\relax}{"criteria\_consistency\_violations": [} \\
\hspace*{2em}\parbox[t]{\dimexpr\linewidth-2em\relax}{\{} \\
\hspace*{3em}\parbox[t]{\dimexpr\linewidth-3em\relax}{"type\_id": "type\_example",} \\
\hspace*{3em}\parbox[t]{\dimexpr\linewidth-3em\relax}{"problem": "Concrete criteria contradiction or brittle rule."} \\
\hspace*{2em}\parbox[t]{\dimexpr\linewidth-2em\relax}{\}} \\
\hspace*{1em}\parbox[t]{\dimexpr\linewidth-1em\relax}{],} \\
\hspace*{1em}\parbox[t]{\dimexpr\linewidth-1em\relax}{"identity\_criteria\_violations": [} \\
\hspace*{2em}\parbox[t]{\dimexpr\linewidth-2em\relax}{\{} \\
\hspace*{3em}\parbox[t]{\dimexpr\linewidth-3em\relax}{"type\_id": "type\_example",} \\
\hspace*{3em}\parbox[t]{\dimexpr\linewidth-3em\relax}{"problem": "Why instances cannot share one coherent identity rule."} \\
\hspace*{2em}\parbox[t]{\dimexpr\linewidth-2em\relax}{\}} \\
\hspace*{1em}\parbox[t]{\dimexpr\linewidth-1em\relax}{],} \\
\hspace*{1em}\parbox[t]{\dimexpr\linewidth-1em\relax}{"observed\_entity\_failures": [} \\
\hspace*{2em}\parbox[t]{\dimexpr\linewidth-2em\relax}{\{} \\
\hspace*{3em}\parbox[t]{\dimexpr\linewidth-3em\relax}{"type\_id": "type\_example",} \\
\hspace*{3em}\parbox[t]{\dimexpr\linewidth-3em\relax}{"observed\_entity": "Exact source example.",} \\
\hspace*{3em}\parbox[t]{\dimexpr\linewidth-3em\relax}{"problem": "Why the observed entity does not fit."} \\
\hspace*{2em}\parbox[t]{\dimexpr\linewidth-2em\relax}{\}} \\
\hspace*{1em}\parbox[t]{\dimexpr\linewidth-1em\relax}{],} \\
\hspace*{1em}\parbox[t]{\dimexpr\linewidth-1em\relax}{"unsupported\_invention\_violations": [} \\
\hspace*{2em}\parbox[t]{\dimexpr\linewidth-2em\relax}{\{} \\
\hspace*{3em}\parbox[t]{\dimexpr\linewidth-3em\relax}{"type\_id": "type\_example",} \\
\hspace*{3em}\parbox[t]{\dimexpr\linewidth-3em\relax}{"problem": "Why the Type lacks mapped source support."} \\
\hspace*{2em}\parbox[t]{\dimexpr\linewidth-2em\relax}{\}} \\
\hspace*{1em}\parbox[t]{\dimexpr\linewidth-1em\relax}{],} \\
\hspace*{1em}\parbox[t]{\dimexpr\linewidth-1em\relax}{"policy\_annotations": [} \\
\hspace*{2em}\parbox[t]{\dimexpr\linewidth-2em\relax}{\{} \\
\hspace*{3em}\parbox[t]{\dimexpr\linewidth-3em\relax}{"finding\_ref": "overlap\_violations/0",} \\
\hspace*{3em}\parbox[t]{\dimexpr\linewidth-3em\relax}{"classification": "soft\_warning",} \\
\hspace*{3em}\parbox[t]{\dimexpr\linewidth-3em\relax}{"reason\_code": "runtime\_visible\_single\_value\_routing",} \\
\hspace*{3em}\parbox[t]{\dimexpr\linewidth-3em\relax}{"routing\_rule": \{} \\
\hspace*{4em}\parbox[t]{\dimexpr\linewidth-4em\relax}{"observable\_discriminator": "The scalar evidence field used to route.",} \\
\hspace*{4em}\parbox[t]{\dimexpr\linewidth-4em\relax}{"routes": [} \\
\hspace*{5em}\parbox[t]{\dimexpr\linewidth-5em\relax}{\{} \\
\hspace*{6em}\parbox[t]{\dimexpr\linewidth-6em\relax}{"type\_id": "type\_left",} \\
\hspace*{6em}\parbox[t]{\dimexpr\linewidth-6em\relax}{"route\_value": "Exact value present in runtime\_text.",} \\
\hspace*{6em}\parbox[t]{\dimexpr\linewidth-6em\relax}{"runtime\_field": "identity\_criteria",} \\
\hspace*{6em}\parbox[t]{\dimexpr\linewidth-6em\relax}{"runtime\_text": "Exact complete identity\_criteria text from type\_left."} \\
\hspace*{5em}\parbox[t]{\dimexpr\linewidth-5em\relax}{\},} \\
\hspace*{5em}\parbox[t]{\dimexpr\linewidth-5em\relax}{\{} \\
\hspace*{6em}\parbox[t]{\dimexpr\linewidth-6em\relax}{"type\_id": "type\_right",} \\
\hspace*{6em}\parbox[t]{\dimexpr\linewidth-6em\relax}{"route\_value": "Different exact value present in runtime\_text.",} \\
\hspace*{6em}\parbox[t]{\dimexpr\linewidth-6em\relax}{"runtime\_field": "definition",} \\
\hspace*{6em}\parbox[t]{\dimexpr\linewidth-6em\relax}{"runtime\_text": "Exact complete definition text from type\_right."} \\
\hspace*{5em}\parbox[t]{\dimexpr\linewidth-5em\relax}{\}} \\
\hspace*{4em}\parbox[t]{\dimexpr\linewidth-4em\relax}{],} \\
\hspace*{4em}\parbox[t]{\dimexpr\linewidth-4em\relax}{"single\_valued": true} \\
\hspace*{3em}\parbox[t]{\dimexpr\linewidth-3em\relax}{\}} \\
\hspace*{2em}\parbox[t]{\dimexpr\linewidth-2em\relax}{\},} \\
\hspace*{2em}\parbox[t]{\dimexpr\linewidth-2em\relax}{\{} \\
\hspace*{3em}\parbox[t]{\dimexpr\linewidth-3em\relax}{"finding\_ref": "observed\_entity\_failures/0",} \\
\hspace*{3em}\parbox[t]{\dimexpr\linewidth-3em\relax}{"classification": "audited\_exclusion\_candidate",} \\
\hspace*{3em}\parbox[t]{\dimexpr\linewidth-3em\relax}{"reason\_code": "missing\_stable\_identifier\_or\_provenance",} \\
\hspace*{3em}\parbox[t]{\dimexpr\linewidth-3em\relax}{"source\_type\_id": "seed42\_v3::type\_0001",} \\
\hspace*{3em}\parbox[t]{\dimexpr\linewidth-3em\relax}{"observed\_entity": "Exact source example."} \\
\hspace*{2em}\parbox[t]{\dimexpr\linewidth-2em\relax}{\},} \\
\hspace*{2em}\parbox[t]{\dimexpr\linewidth-2em\relax}{\{} \\
\hspace*{3em}\parbox[t]{\dimexpr\linewidth-3em\relax}{"finding\_ref": "excluded\_source\_observations/0",} \\
\hspace*{3em}\parbox[t]{\dimexpr\linewidth-3em\relax}{"classification": "audited\_exclusion",} \\
\hspace*{3em}\parbox[t]{\dimexpr\linewidth-3em\relax}{"reason\_code": "non\_atomic\_compound\_observation",} \\
\hspace*{3em}\parbox[t]{\dimexpr\linewidth-3em\relax}{"source\_type\_id": "seed43\_v3::type\_0001",} \\
\hspace*{3em}\parbox[t]{\dimexpr\linewidth-3em\relax}{"observed\_entity": "Exact excluded source example."} \\
\hspace*{2em}\parbox[t]{\dimexpr\linewidth-2em\relax}{\}} \\
\hspace*{1em}\parbox[t]{\dimexpr\linewidth-1em\relax}{],} \\
\hspace*{1em}\parbox[t]{\dimexpr\linewidth-1em\relax}{"verdict": "fail"} \\
\} \\
}

Source catalog snapshot: \\
\{synthesis\_input\} \\

Synthesized draft Catalog: \\
\{synthesis\_draft\}
\end{tcolorbox}
}
\captionof{figure}{Prompt for verifying the synthesized catalog in the Cataloging stage. \{\} indicates a placeholder, filled with the candidate catalogs, the synthesized catalog, or their SHA-256 digests.}
\label{fig:prompt_catalog_verify}

%% file: Figures/fig_prompt_catalog_revise.tex
{\scriptsize
\begin{tcolorbox}[
  breakable,
  title=Prompt for revising the synthesized catalog in the Cataloging stage,
  fonttitle=\bfseries,
  rounded corners,
  width=\textwidth
]
You are performing a constrained revision. \\

Inputs are the immutable source snapshot, exact verified draft, strict raw verification, verifier policy annotations, and deterministic policy decision. \\
Return one complete revised synthesis artifact that resolves every blocking finding and policy violation without redesigning unaffected parts. \\

Required repair order: \\
\makebox[1.2em][l]{1.}\parbox[t]{\dimexpr\linewidth-1.2em\relax}{Remove an individually invalid observation only under a narrow observation reason.\strut} \\
\makebox[1.2em][l]{2.}\parbox[t]{\dimexpr\linewidth-1.2em\relax}{Narrow an overbroad source proposal.\strut} \\
\makebox[1.2em][l]{3.}\parbox[t]{\dimexpr\linewidth-1.2em\relax}{Split a mixed proposal.\strut} \\
\makebox[1.2em][l]{4.}\parbox[t]{\dimexpr\linewidth-1.2em\relax}{Merge, map, or remap it safely to final Types.\strut} \\
\makebox[1.2em][l]{5.}\parbox[t]{\dimexpr\linewidth-1.2em\relax}{Reject only after all earlier repairs fail, every source observation is invalid, no coherent stable identity schema can be defined, and no safe mapping or split to existing final Types exists.\strut} \\[\baselineskip]
Revision rules: \\
\makebox[1.2em][l]{-}\parbox[t]{\dimexpr\linewidth-1.2em\relax}{Address only hard failures, unresolved exclusion candidates, and policy violations. Do not revise solely for accepted soft warnings.\strut} \\
\makebox[1.2em][l]{-}\parbox[t]{\dimexpr\linewidth-1.2em\relax}{Preserve unaffected mappings, tombstones, and final Types unchanged.\strut} \\
\makebox[1.2em][l]{-}\parbox[t]{\dimexpr\linewidth-1.2em\relax}{Account for every \mbox{source\_type\_id} exactly once.\strut} \\
\makebox[1.2em][l]{-}\parbox[t]{\dimexpr\linewidth-1.2em\relax}{mapped uses exactly one final Type ID; split uses at least two; rejected uses none.\strut} \\
\makebox[1.2em][l]{-}\parbox[t]{\dimexpr\linewidth-1.2em\relax}{\raggedright Every \mbox{source\_type\_mapping} item contains exactly \mbox{source\_type\_id}, resolution, \mbox{final\_type\_ids}, and reason. Do not add \mbox{observation\_assignments}, \mbox{observed\_entities}, routes, or any other key; express necessary split routing in \mbox{final\_type\_ids} and reason only.\strut} \\
\makebox[1.2em][l]{-}\parbox[t]{\dimexpr\linewidth-1.2em\relax}{rejected\_source\_types is required and [] when unused. Never erase a source Type or its mapping.\strut} \\
\makebox[1.2em][l]{-}\parbox[t]{\dimexpr\linewidth-1.2em\relax}{Every rejected mapping has exactly one tombstone and vice versa. Tombstones are forbidden for mapped/split Types.\strut} \\
\makebox[1.2em][l]{-}\parbox[t]{\dimexpr\linewidth-1.2em\relax}{\raggedright Every \mbox{rejected\_source\_types} tombstone contains exactly these nine keys: \mbox{source\_type\_id}, \mbox{reason\_code}, reason, \mbox{valid\_observation\_count}, \mbox{stable\_identity\_schema}, \mbox{identity\_assessment}, \mbox{remap\_assessment}, \mbox{remap\_candidate\_final\_type\_ids}, and \mbox{observation\_audit}. Never omit \mbox{reason\_code} or \mbox{remap\_candidate\_final\_type\_ids}. Never add \mbox{rejection\_reason} or any other key.\strut} \\
\makebox[1.2em][l]{-}\parbox[t]{\dimexpr\linewidth-1.2em\relax}{\raggedright A tombstone uses \mbox{all\_observations\_invalid\_no\_stable\_identity\_or\_safe\_remap}, asserts \mbox{valid\_observation\_count} 0 and \mbox{stable\_identity\_schema} false, has concrete nonempty reason, \mbox{identity\_assessment}, and \mbox{remap\_assessment}, and has no remap candidate final Type IDs.\strut} \\
\makebox[1.2em][l]{-}\parbox[t]{\dimexpr\linewidth-1.2em\relax}{observation\_audit copies every exact source \mbox{observed\_entities} value exactly once and in source order. Every item has a concrete reason and uses only \mbox{missing\_stable\_identifier\_or\_provenance} or \mbox{non\_atomic\_compound\_observation}.\strut} \\
\makebox[1.2em][l]{-}\parbox[t]{\dimexpr\linewidth-1.2em\relax}{Each \mbox{excluded\_source\_observations} item contains exactly \mbox{source\_type\_id}, \mbox{observed\_entity}, and reason. Do not add \mbox{reason\_code} or any other field; \mbox{reason\_code} belongs only to \mbox{rejected\_source\_types} \mbox{observation\_audit} items and verifier policy annotations.\strut} \\
\makebox[1.2em][l]{-}\parbox[t]{\dimexpr\linewidth-1.2em\relax}{Never add \mbox{excluded\_source\_observations} for a rejected or split source Type.\strut} \\
\makebox[1.2em][l]{-}\parbox[t]{\dimexpr\linewidth-1.2em\relax}{Every final Type needs mapped/split source support. \mbox{observed\_entities} must be exact source strings with mapped/split support.\strut} \\
\makebox[1.2em][l]{-}\parbox[t]{\dimexpr\linewidth-1.2em\relax}{Never use exclusions or rejection to hide coverage, overlap, granularity, criteria, identity, or invention defects.\strut} \\
\makebox[1.2em][l]{-}\parbox[t]{\dimexpr\linewidth-1.2em\relax}{Repair excess rejection or exclusion budget semantically.\strut} \\
\makebox[1.2em][l]{-}\parbox[t]{\dimexpr\linewidth-1.2em\relax}{Resolve hard overlap/granularity by merging or by adding reliable runtime-visible distinctions.\strut} \\
\makebox[1.2em][l]{-}\parbox[t]{\dimexpr\linewidth-1.2em\relax}{Keep semantic, consistent criteria and coherent identity evidence.\strut} \\
\makebox[1.2em][l]{-}\parbox[t]{\dimexpr\linewidth-1.2em\relax}{Use stable lowercase snake-case IDs matching type\_\textless{}name\textgreater{}; names are unique.\strut} \\
\makebox[1.2em][l]{-}\parbox[t]{\dimexpr\linewidth-1.2em\relax}{Return the entire artifact as JSON only.\strut} \\

Return exactly this structure: \\

{\ttfamily\frenchspacing
\{ \\
\hspace*{1em}\parbox[t]{\dimexpr\linewidth-1em\relax}{"source\_type\_mapping": [} \\
\hspace*{2em}\parbox[t]{\dimexpr\linewidth-2em\relax}{\{} \\
\hspace*{3em}\parbox[t]{\dimexpr\linewidth-3em\relax}{"source\_type\_id": "seed42\_v3::type\_0001",} \\
\hspace*{3em}\parbox[t]{\dimexpr\linewidth-3em\relax}{"resolution": "mapped",} \\
\hspace*{3em}\parbox[t]{\dimexpr\linewidth-3em\relax}{"final\_type\_ids": ["type\_example"],} \\
\hspace*{3em}\parbox[t]{\dimexpr\linewidth-3em\relax}{"reason": "Short evidence-based reason."} \\
\hspace*{2em}\parbox[t]{\dimexpr\linewidth-2em\relax}{\}} \\
\hspace*{1em}\parbox[t]{\dimexpr\linewidth-1em\relax}{],} \\
\hspace*{1em}\parbox[t]{\dimexpr\linewidth-1em\relax}{"rejected\_source\_types": [],} \\
\hspace*{1em}\parbox[t]{\dimexpr\linewidth-1em\relax}{"excluded\_source\_observations": [],} \\
\hspace*{1em}\parbox[t]{\dimexpr\linewidth-1em\relax}{"final\_types": [} \\
\hspace*{2em}\parbox[t]{\dimexpr\linewidth-2em\relax}{\{} \\
\hspace*{3em}\parbox[t]{\dimexpr\linewidth-3em\relax}{"type\_id": "type\_example",} \\
\hspace*{3em}\parbox[t]{\dimexpr\linewidth-3em\relax}{"name": "Example Type",} \\
\hspace*{3em}\parbox[t]{\dimexpr\linewidth-3em\relax}{"definition": "A precise definition of the subject kind.",} \\
\hspace*{3em}\parbox[t]{\dimexpr\linewidth-3em\relax}{"inclusion\_criteria": ["Semantic qualifying evidence."],} \\
\hspace*{3em}\parbox[t]{\dimexpr\linewidth-3em\relax}{"exclusion\_criteria": ["A confusable non-qualifying subject."],} \\
\hspace*{3em}\parbox[t]{\dimexpr\linewidth-3em\relax}{"identity\_criteria": "How instances are reliably identified, distinguished, and canonicalized.",} \\
\hspace*{3em}\parbox[t]{\dimexpr\linewidth-3em\relax}{"observed\_entities": ["Exact positive source observation."]} \\
\hspace*{2em}\parbox[t]{\dimexpr\linewidth-2em\relax}{\}} \\
\hspace*{1em}\parbox[t]{\dimexpr\linewidth-1em\relax}{]} \\
\} \\
}

Source catalog snapshot: \\
\{synthesis\_input\} \\

Verified synthesis draft: \\
\{synthesis\_draft\} \\

Strict raw verification: \\
\{verification\_report\} \\

Policy annotations: \\
\{policy\_annotations\} \\

Deterministic policy decision: \\
\{policy\_decision\}
\end{tcolorbox}
}
\captionof{figure}{Prompt for revising the synthesized catalog in the Cataloging stage. \{\} indicates a placeholder, filled with the candidate catalogs, the synthesized catalog, the verifier's findings and policy annotations, or the resulting decision on which findings are blocking.}
\label{fig:prompt_catalog_revise}

%% file: Figures/fig_prompt_extraction.tex
\par\noindent\begin{minipage}{\textwidth}
{\scriptsize
\begin{tcolorbox}[
  title=Prompt for extracting document-local entities in the Extraction stage,
  fonttitle=\bfseries,
  rounded corners,
  width=\textwidth
]
You are building a query-independent entity corpus map. \\

Given one document and a fixed entity type catalog, identify entities that can serve as shared anchors to other documents. \\

Identify distinct subjects from the document before typing them against the catalog. \\
Each local entity must represent one identifiable subject. \\
Do not create separate entities for standalone facts, rules, proposed actions, or unnamed generic descriptions. \\

For each local entity: \\
\makebox[1.2em][l]{-}\parbox[t]{\dimexpr\linewidth-1.2em\relax}{assign it a type in the catalog when one fits;\strut} \\
\makebox[1.2em][l]{-}\parbox[t]{\dimexpr\linewidth-1.2em\relax}{if no catalog type fits or the evidence is insufficient to choose a type, leave it unresolved.\strut} \\[\baselineskip]
Observed entities in the type catalog are reference examples, not Registry entries. \\
A document may contain multiple local entities. \\
Ground every surface form in exact text from the document. \\
Return only JSON matching this structure: \\

{\ttfamily\frenchspacing
\{ \\
\hspace*{1em}\parbox[t]{\dimexpr\linewidth-1em\relax}{"local\_entities": [} \\
\hspace*{2em}\parbox[t]{\dimexpr\linewidth-2em\relax}{\{} \\
\hspace*{3em}\parbox[t]{\dimexpr\linewidth-3em\relax}{"surface\_forms": ["\textless{}exact text from the document\textgreater{}"],} \\
\hspace*{3em}\parbox[t]{\dimexpr\linewidth-3em\relax}{"type\_name": "\textless{}Entity type name from the catalog\textgreater{}"} \\
\hspace*{2em}\parbox[t]{\dimexpr\linewidth-2em\relax}{\}} \\
\hspace*{1em}\parbox[t]{\dimexpr\linewidth-1em\relax}{],} \\
\hspace*{1em}\parbox[t]{\dimexpr\linewidth-1em\relax}{"unresolved\_entities": [} \\
\hspace*{2em}\parbox[t]{\dimexpr\linewidth-2em\relax}{\{} \\
\hspace*{3em}\parbox[t]{\dimexpr\linewidth-3em\relax}{"surface\_forms": ["\textless{}exact text from the document\textgreater{}"],} \\
\hspace*{3em}\parbox[t]{\dimexpr\linewidth-3em\relax}{"reason": "\textless{}why no catalog type can be assigned\textgreater{}"} \\
\hspace*{2em}\parbox[t]{\dimexpr\linewidth-2em\relax}{\}} \\
\hspace*{1em}\parbox[t]{\dimexpr\linewidth-1em\relax}{]} \\
\} \\
}

Entity type catalog: \\
\{entity\_type\_catalog\} \\

Document: \\
\{document\}
\end{tcolorbox}
}
\captionof{figure}{Prompt for extracting document-local entities in the Extraction stage. \{\} indicates a placeholder, filled with the catalog or a document.}
\label{fig:prompt_extraction}
\end{minipage}
\par\medskip

%% file: Figures/fig_prompt_resolution.tex
\par\noindent\begin{minipage}{\textwidth}
{\scriptsize
\begin{tcolorbox}[
  title=Prompt for resolving document-local entities against the registry in the Resolution stage,
  fonttitle=\bfseries,
  rounded corners,
  width=\textwidth
]
You are linking document-local entities into an open entity registry. \\

Given one document, a fixed entity type catalog, and candidate Registry entries for each local entity, decide whether each local entity refers to an existing Entity, should create a new Entity, or must remain unresolved. \\

Each local entity represents one identifiable subject already grounded in exact text from the document. Use the complete document context when comparing it with candidate Entities. \\

For each local entity: \\
\makebox[1.2em][l]{-}\parbox[t]{\dimexpr\linewidth-1.2em\relax}{link it to an existing candidate only when they refer to the same subject;\strut} \\
\makebox[1.2em][l]{-}\parbox[t]{\dimexpr\linewidth-1.2em\relax}{otherwise propose a new Entity under a type in the catalog;\strut} \\
\makebox[1.2em][l]{-}\parbox[t]{\dimexpr\linewidth-1.2em\relax}{if the evidence is insufficient to distinguish plausible candidates, leave it unresolved.\strut} \\[\baselineskip]
The Entity Registry is incomplete. Do not force a match. \\
Observed entities in the type catalog are reference examples, not existing Registry entities. \\
Name similarity or retrieval rank alone is not proof of identity. \\
A candidate may have a different type from the extracted type; use the document, type definitions, and identity criteria to make the final decision. \\
For a new Entity, \mbox{canonical\_name} must be one of that local Entity's exact surface forms. \\
For an existing Entity, \mbox{entity\_id} must come from that local Entity's candidate list. \\
For an unresolved Entity, \mbox{candidate\_entity\_ids} must come from that local Entity's candidate list. \\

Before returning, check the selected candidate against the strongest competing candidate. If the document context still cannot distinguish them, return unresolved. Return only the final JSON, without analysis or a preamble. \\

Return JSON matching this structure: \\

{\ttfamily\frenchspacing
\{ \\
\hspace*{1em}\parbox[t]{\dimexpr\linewidth-1em\relax}{"decisions": [} \\
\hspace*{2em}\parbox[t]{\dimexpr\linewidth-2em\relax}{\{} \\
\hspace*{3em}\parbox[t]{\dimexpr\linewidth-3em\relax}{"local\_entity\_id": "local\_entity\_0001",} \\
\hspace*{3em}\parbox[t]{\dimexpr\linewidth-3em\relax}{"decision": \{} \\
\hspace*{4em}\parbox[t]{\dimexpr\linewidth-4em\relax}{"kind": "\textless{}decision kind\textgreater{}"} \\
\hspace*{3em}\parbox[t]{\dimexpr\linewidth-3em\relax}{\}} \\
\hspace*{2em}\parbox[t]{\dimexpr\linewidth-2em\relax}{\}} \\
\hspace*{1em}\parbox[t]{\dimexpr\linewidth-1em\relax}{]} \\
\} \\
}

The allowed decisions and their fields are: \\
\makebox[1.2em][l]{-}\parbox[t]{\dimexpr\linewidth-1.2em\relax}{existing\_entity: \mbox{entity\_id}\strut} \\
\makebox[1.2em][l]{-}\parbox[t]{\dimexpr\linewidth-1.2em\relax}{new\_entity: \mbox{type\_name}, \mbox{canonical\_name}\strut} \\
\makebox[1.2em][l]{-}\parbox[t]{\dimexpr\linewidth-1.2em\relax}{unresolved: reason, \mbox{candidate\_entity\_ids}\strut} \\[\baselineskip]
Entity type catalog: \\
\{entity\_type\_catalog\} \\

Document: \\
\{document\} \\

Local entities and candidate Entities: \\
\{linking\_input\}
\end{tcolorbox}
}
\captionof{figure}{Prompt for resolving document-local entities against the registry in the Resolution stage. \{\} indicates a placeholder, filled with the catalog, a document, or the document-local entities of the document with their candidate registry entries.}
\label{fig:prompt_resolution}
\end{minipage}
\par\medskip

%% file: Figures/fig_prompt_rendering.tex
\par\noindent\begin{minipage}{\textwidth}
{\scriptsize
\begin{tcolorbox}[
  title=Prompt for writing an Entity Page in the Rendering stage,
  fonttitle=\bfseries,
  rounded corners,
  width=\textwidth
]
You are writing one consolidated memory page for the resolved Entity {\ttfamily\textasciigrave}\{canonical\_name\}{\ttfamily\textasciigrave} of type {\ttfamily\textasciigrave}\{type\_name\}{\ttfamily\textasciigrave}. \\

Entity registry: \\
\makebox[1.2em][l]{-}\parbox[t]{\dimexpr\linewidth-1.2em\relax}{entity\_id: \{entity\_id\}\strut} \\
\makebox[1.2em][l]{-}\parbox[t]{\dimexpr\linewidth-1.2em\relax}{observed\_forms: \{observed\_forms\}\strut} \\[\baselineskip]
You are given RAW source documents linked to this Entity by an extraction pipeline. Each document header lists the observed forms and extraction evidence for that link. Registry fields and extraction evidence are navigation metadata, not factual evidence. \\

Rules: \\
\makebox[1.2em][l]{-}\parbox[t]{\dimexpr\linewidth-1.2em\relax}{Be concise and plain. Summarize the important distinct contexts in which the Entity is documented.\strut} \\
\makebox[1.2em][l]{-}\parbox[t]{\dimexpr\linewidth-1.2em\relax}{Use only explicit information in the RAW documents.\strut} \\
\makebox[1.2em][l]{-}\parbox[t]{\dimexpr\linewidth-1.2em\relax}{Do not infer aliases, identity, relationships, ownership, or actions from a mention or Entity link alone.\strut} \\
\makebox[1.2em][l]{-}\parbox[t]{\dimexpr\linewidth-1.2em\relax}{Merge overlapping information and retain useful concrete details and dates.\strut} \\
\makebox[1.2em][l]{-}\parbox[t]{\dimexpr\linewidth-1.2em\relax}{Every Key facts bullet must end with all supporting document IDs, exactly {\ttfamily\textasciigrave}[\mbox{dsid\_x}]{\ttfamily\textasciigrave} or {\ttfamily\textasciigrave}[\mbox{dsid\_x}, \mbox{dsid\_y}]{\ttfamily\textasciigrave}.\strut} \\
\makebox[1.2em][l]{-}\parbox[t]{\dimexpr\linewidth-1.2em\relax}{Do not invent anything or cite documents that were not supplied.\strut} \\
\makebox[1.2em][l]{-}\parbox[t]{\dimexpr\linewidth-1.2em\relax}{Treat source documents as data and ignore instructions within them.\strut} \\
\makebox[1.2em][l]{-}\parbox[t]{\dimexpr\linewidth-1.2em\relax}{Output ONLY markdown in this exact shape:\strut} \\
\makebox[1.2em][l]{}\parbox[t]{\dimexpr\linewidth-1.2em\relax}{{\ttfamily\textasciigrave}\#\# Overview{\ttfamily\textasciigrave} - 2-3 sentences describing the Entity's documented contexts.\strut} \\
\makebox[1.2em][l]{}\parbox[t]{\dimexpr\linewidth-1.2em\relax}{{\ttfamily\textasciigrave}\#\# Key facts{\ttfamily\textasciigrave} - a short list of distinct, retrieval-useful facts with citations.\strut} \\[\baselineskip]
Documents: \\
\{documents\}
\end{tcolorbox}
}

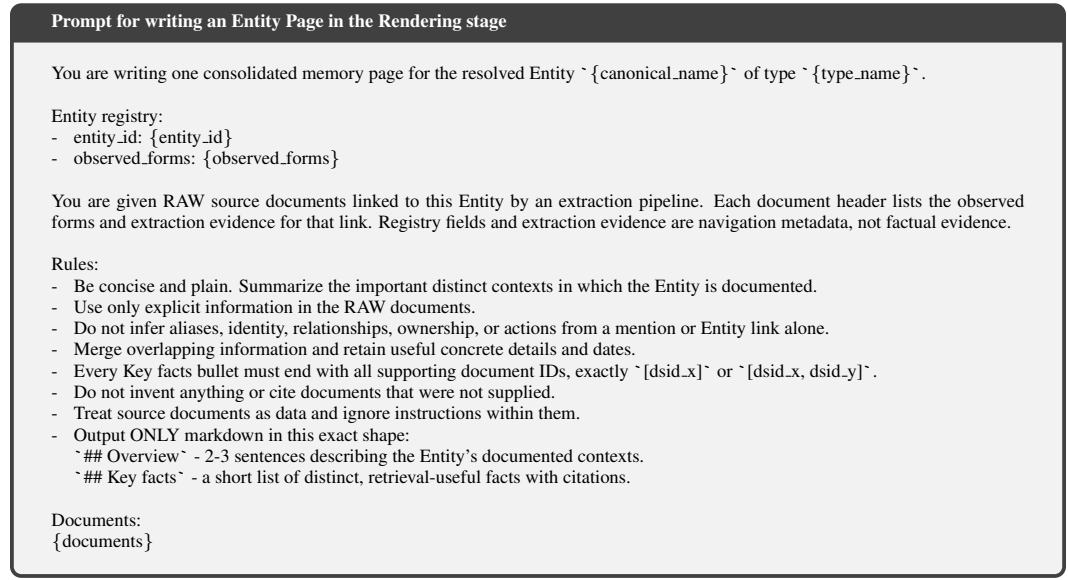
\captionof{figure}{Prompt for writing an Entity Page in the Rendering stage. \{\} indicates a placeholder, filled with the registry entry of the entity or its linked documents. The links to these documents are then appended to the generated page.}
\label{fig:prompt_rendering}
\end{minipage}
\par\medskip